\documentclass{article}
\usepackage{ijcai22}

\usepackage{times}
\usepackage{soul}
\usepackage{url}
\usepackage[hidelinks]{hyperref}
\usepackage[utf8]{inputenc}
\usepackage[small]{caption}
\usepackage{graphicx}
\usepackage{amsmath}
\usepackage{amsthm}
\usepackage{booktabs}
\usepackage{algorithm}
\usepackage{algorithmic}
\usepackage[title]{appendix}
\usepackage{algorithm}
\usepackage{algorithmic}
\usepackage{xcolor}
\usepackage{multirow}
\usepackage{amsmath}
\usepackage{pifont}
\usepackage[capitalize]{cleveref}
\newcommand{\cmark}{\ding{51}}%

\newcommand{\etal}{\textit{et al.}}

\title{Mutual Guidance and Residual Integration for Image Enhancement}

	\author{
	Kun Zhou\textsuperscript{1,2},
	\quad KenKun Liu\textsuperscript{1,2},
	\quad Wenbo Li\textsuperscript{3},
	\quad Xiaoguang Han\textsuperscript{1},
	\quad Jiangbo Lu\textsuperscript{2}% <-this % stops a space
	\affiliations
	\quad ${^1}$The Chinese University of Hong Kong~(Shenzhen) $^{2}$SmartMore Corporation \\ 
	$^{3}$The Chinese University of Hong Kong\\
}

\begin{document}
\maketitle

\begin{abstract}

	Previous studies show the necessity of global and local adjustment for image enhancement. However, existing convolutional neural networks~(CNNs) and transformer-based models face great challenges in balancing computational efficiency and effectiveness of global-local information usage. Especially, existing methods typically adopt the global-to-local fusion mode, ignoring the importance of bidirectional interactions.  To address those issues, we propose a novel mutual guidance network~(MGN) to perform effective bidirectional global-local information exchange while keeping a compact architecture. In our design, we adopt a two-branch framework where one branch focuses more on modeling global relations while the other is committed to processing local information. Then, we develop an efficient attention-based mutual guidance approach throughout our framework for bidirectional global-local interactions. As a result, both the global and local branches can enjoy the merits of mutual information aggregation. Besides, to further refine the results produced by our MGN, we propose a novel residual integration scheme following the divide-and-conquer philosophy. The extensive experiments demonstrate the effectiveness of our proposed method, which achieves state-of-the-art performance on several public image enhancement benchmarks.
	
\end{abstract}

\section{Introduction}
% histogram equalization~(HE)~\cite{pizer1990contrast} or gamma correction~\cite{guan2009image}
Low-quality image enhancement is a challenging problem since it not only needs to adjust the brightness, local contrast, and color but also requires to maintain global contextual consistency. 
On the one hand, methods for global enhancement typically employ uniform transforms, e.g., histogram equalization (HE)~\cite{pizer1990contrast} or gamma
correction~\cite{guan2009image} yet failing to adaptively recover detailed image contents. On the other hand, the line of local enhancement focuses on adjusting the brightness or contrast within image patches, sometimes producing inconsistent results. Therefore, it is critical to perform effective global and local information fusion.
%	\textcolor{red} { Global enhancement utilize limited global transforms for fast adjustment, e.g., histogram equalization (HE)~\cite{pizer1990contrast} or gamma
%	correction~\cite{guan2009image}. On the other hand, local enhancement methods focus on maximizing the brightness/contrast within image patches. However, the former fails to recover detailed image contents and the later suffers from the inconsistent enhancement between image patches and produces hole artifacts. Therefore, it is critical to perform effective global and local information aggregation.}
In the view of global-local content fusion, extensive approaches have been proposed. For example, HDRNet~\cite{gharbi2017deep} presents a global-local structure that utilizes an average-pooling operation to capture global information
and achieve significant improvements over the previous approaches~\cite{hu2018exposure,park2018distort}. Some follow-up works~\cite{Wang_2019_CVPR,chen2018deep} adopt a similar design for capturing global content. However, the inherent properties of CNNs show the limited ability for global content extraction.

In contrast to CNN-based neural networks, recent works~\cite{xu2022snr,wang2022structural,zhang2021star,cui2022illumination,kim2021representative,peng2021u,souibgui2022docentr} begin to exploit various transformer architectures for image enhancement. Benefiting from the attention mechanism, these methods are able to model long-range information for better global consistency. But, such designs require high computational costs and prohibit generalizing to high-resolution inputs.  
% RCT~\cite{kim2021representative} uses a vision transformer to learns sparse representative color calibration. Through it achieves global enhancement, but the local contrast can not be guaranteed. To obtain  global-local image information, Wang~\etal~\cite{wang2022structural} proposes a U-Shape transformer that adopts stacked window-based attention blocks for image enhancement. As aforementioned, it suffers from highly computational burdens and prohibits generalizing high-resolution inputs. 

\begin{figure*}[t]
	\centering
	\includegraphics[width=1.0\linewidth]{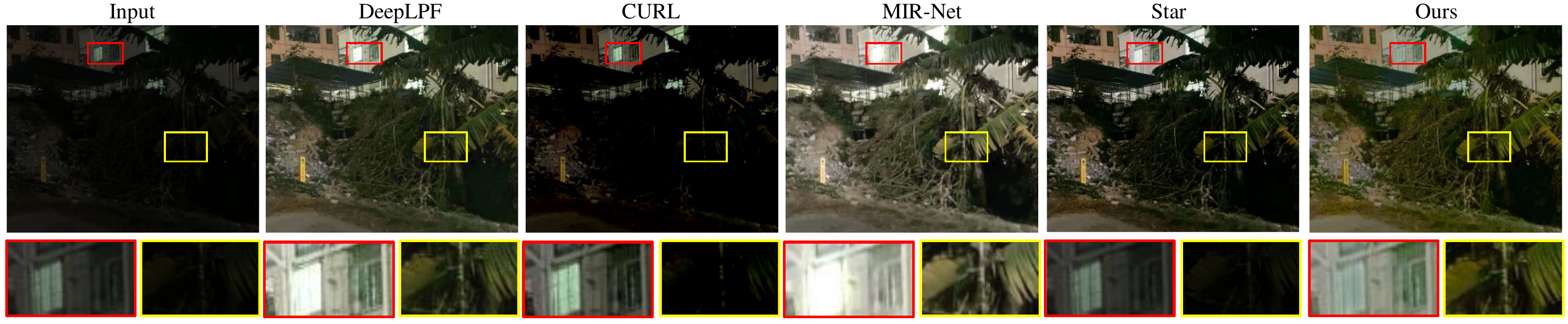} 
	\caption{A low-light image captured by a mobile phone. Compared with SOTA methods, our model correctly brightens dark areas and restores pleasing contrasts.  } 
	\label{fig:teasing}
\end{figure*}

We find that both the existing CNNs~\cite{hu2018exposure,park2018distort,Wang_2019_CVPR,chen2018deep} that employ the channel concatenation strategy and the transformer models~\cite {wang2022structural,cui2022illumination,peng2021u,souibgui2022docentr} that stack multiple attention blocks have not considered bidirectional global-local fusion. Without adequate mutual global-local interactions, as illustrated in Figure~\ref{fig:teasing}, these methods still suffer from under-/over exposure issues.
\begin{figure}[t]
	\centering
	\includegraphics[width=1.0\columnwidth]{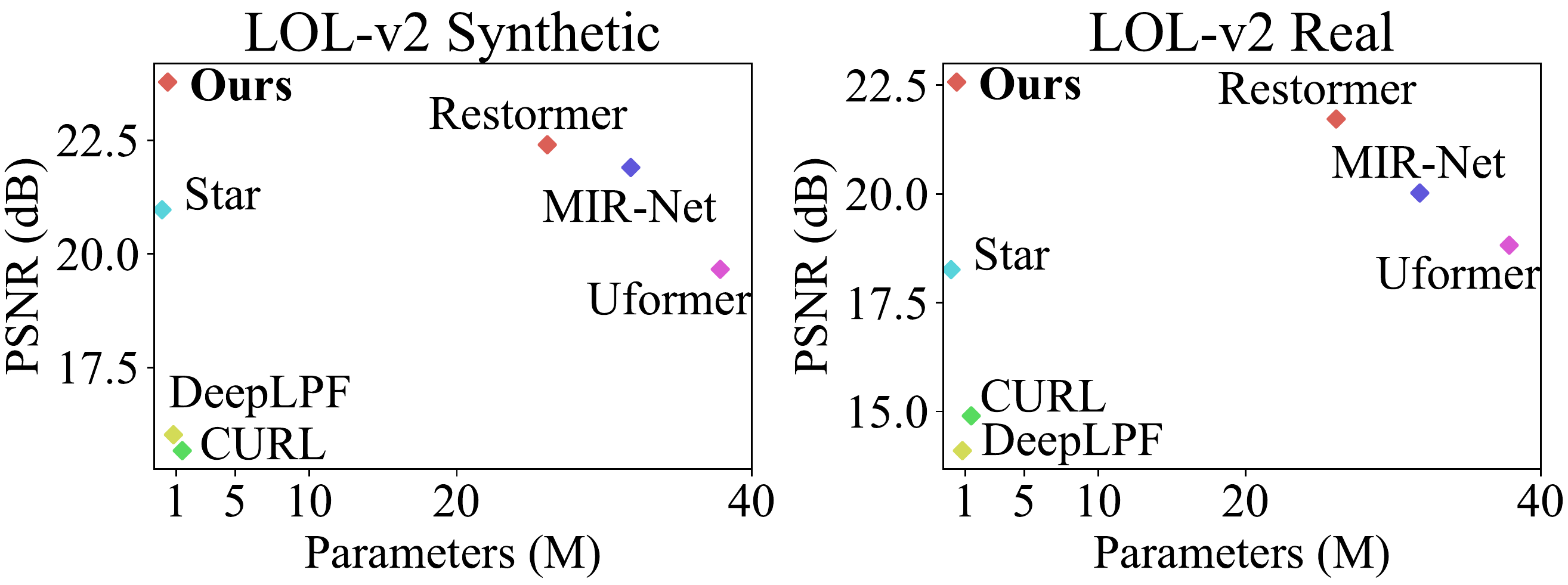} 
	\caption{Comparison between our proposed model and other approaches on LOL-v2 Synthetic and LOL-v2 Real. } 
	\label{fig:params}
	
\end{figure}

To address this issue, we develop a novel mutual guidance network~(dubbed MGN) for image enhancement. 
The proposed MGN is composed of two parallel branches~(namely global and local branches) that are interconnected with each other. We first utilize a transformer head and a CNN head for extracting global and local features, respectively.
To ensure effective global-local fusion, we design a mutual guidance structure that performs bidirectional information exchange between the two branches. Instead of one-shot mutual aggregation, we develop novel cascading mutual guidance blocks for effective global-local fusion, realizing high-frequency mutual interaction. Meanwhile, we devise a channel-wise co-attention mechanism renders the fusion extremely efficient, largely reducing the computational cost compared with the conventional self-attention or sliced window attention designs. Thanks to the mutual guidance modeling, our MGN is able to generate image residues with rich global-local enhancement contents. 
Based on the informative residues, we further formulate the image enhancement as a residual integration process. Following the divide-conquer strategy, we coarse-to-finely divide the enhancement residues into several residual pieces and learn importance maps for piece-wise integration. With the proposed designs, our model achieves more accurate enhancement.  As shown in Figure~\ref{fig:teasing}, compared with other SOTA methods~\cite{moran2020deeplpf,moran2021curl,Zamir2020MIRNet,zamir2022restormer}, our approach generates the most visually pleasant results in view of local contrast, natural color and global consistency.

In total, the contributions of this paper are three-fold.
\begin{itemize}
	\item We develop an effective mutual guidance network for image enhancement. We present a novel efficient attention mechanism~(${\mathcal O}(C)$ computational complexity where $C$ is the channel dimension\footnote{In the low-level vision, channel dimension is quite smaller than the image resolution~($C \ll H*W$)}.) for bidirectional global and local information exchange. Empirical experiments show our MGN is capable of enhancing low-quality images with better brightness and local contrast, while maintaining global consistency.
	
	\item We introduce a residual integration scheme to refine the residues produced by our MGN. By integrating piece-wise residual information with respect to its predicted importance, it is more flexible and effective for image enhancement. 
	\item Thanks to our mutual guidance aggregation and integral residual learning, our model achieves state-of-the-art performance on various image enhancement benchmarks.
\end{itemize}

\section{Related Works}

\subsection{CNN-Based Methods}
Recently, the CNN-based low-quality image enhancement has been extensively studied~\cite{xu2020learning,moran2020deeplpf,shen2017msr,yang2016enhancement,Wang_2019_CVPR}.
Yang \etal~\cite{yang2016enhancement} enhance a low-light image using coupled dictionary learning.
DeepLPF~\cite{moran2020deeplpf} employs a deep neural network to regress three types of local parametric filters for automatic enhancement. GleNet~\cite{kim2020global} proposes a global-local enhancement framework for image enhancement. HDRNet~\cite{gharbi2017deep} introduces an averaging pooling operation to capture global contextual information and achieves significant improvement over previous approaches~\cite{hu2018exposure,park2018distort}. Encouraged by their works, many studies~\cite{Wang_2019_CVPR,chen2018deep,kinoshita2019convolutional} adopt similar designs for global-local enhancement. However, as aforementioned, CNNs lack the ability of modeling long-range dependencies, leading to inadequate global contextual extraction.
% xu2020learning
%---------------------------------------------------------------------------------------------------------------
\begin{figure*}[t]
	\centering
	\includegraphics[width=1.0\linewidth]{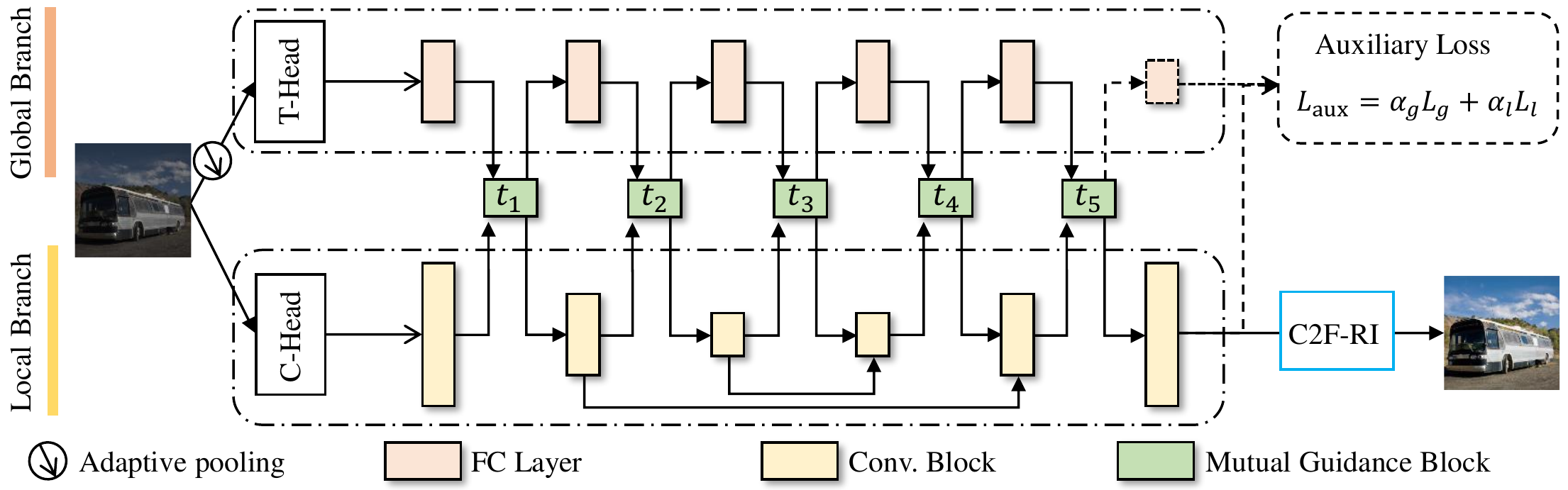} 
	\caption{Our MGN is composed of two interconnected branches. The global and branch utilize a transformer head~(T-head) and a CNN head~(C-head) for feature extraction. Then we stack 5~($t_1,t_2,\cdots,t_5$) mutual guidance blocks for effective global-local feature fusion. After mutual guidance aggregation, the updated global and local features are passed through their respective branches for further processing.  At the end of the two branches, our coarse-to-fine residual integration~(``C2F-RI") block receives the enhancement residue $R_x$ produced by our MGN and employs division-and-integration steps for residue refinement. } 
	\label{fig:framework}
\end{figure*} % 
%---------------------------------------------------------------------------------------------------------------
\subsection{Transformer-based Methods}
Motivated by the great success of transformer, a lot of works~\cite{zheng2021rethinking,chen2021pre,lanchantin2021general,li2021lifting,wang2021end} propose various transformer frameworks to solve different computer vision tasks, e.g., semantic/instance segmentation~\cite{zheng2021rethinking,wang2021end,wang2021max}, human pose estimation~\cite{lin2021end,li2021lifting}. The main advantage of the transformer-based framework is its ability to capture long-range dependencies through attention mechanisms. More recently, RCT~\cite{kim2021representative} uses a vision transformer to regress representative colors and the accordingly transformed color values for image enhancement. Star~\cite{zhang2021star} uses a lightweight transformer to learn coarse image structure information for image enhancement. Uformer~\cite{wang2021uformer} develop a U-Shape transformer to capture both global and local information. Though the attention nature of transformer-based approaches~\cite{xu2022snr,wang2022structural,zhang2021star,cui2022illumination,kim2021representative,peng2021u,zamir2022restormer} enables better global reasoning ability, it is still difficult to simultaneously model global-local image contents, while achieving computational efficiency.

\subsection{Global-Local Fusion}
Feature concatenation is widely employed in CNN architectures. Especially, they embed the global 1D feature into a convolved local feature by sticking them along the channel dimension. Though it is straightforward, the simple combination lacks sufficient global-local information exchange. Therefore, it cannot guarantee the enhancement quality. On the other hand, due to the attention mechanism, transformer-based models show superiority in modeling long-range dependencies. Stacking multiple patch-wise attention blocks, the window-based transformers~\cite{wang2021uformer,wang2022structural,song2021starenhancer} achieves better global-local interactions, thus leading to superior performance than CNNs.
However, the expensive computation costs discourage the application of high-resolution scenarios. In this paper, we present a novel mutual guidance network that explores bidirectional global-local information fusion effectively and efficiently. Besides, we design an integral residual learning algorithm that is able to adaptively aggregate the residues produced by our MGN coarse-to-finely. With the proposed mutual guidance mechanism and residual integration, our model achieves a new state-of-the-art performance over existing image enhancement approaches.

\section{Our Method}\label{sec:method}
In this section, we will first briefly review previous global-local enhancement frameworks. Then, we will introduce our proposed mutual guidance network and our residual integration algorithm, separately. 

\subsection{Global-local Enhancement}\label{sec:preview}
The global-local enhancement approaches can be roughly divided into two classes: (1) CNN-based models; (2) transformer-based methods. For CNN-based models, they obtain the 1D global feature by global average poolings. As aforementioned, such an averaging operation is less effective to capture long-range dependencies between all image pixel pairs. For transformer-based methods, to obtain both global and local image information, they require to stack multiple self-attention blocks that calculate window-wise correlations~\cite{wang2021uformer,wang2022structural,song2021starenhancer}. It leads to high computational complexity and prohibits generalizing to high-resolution scenarios.

\subsection{Mutual Guidance Enhancement}
To address those issues as well as for better global-local interaction, we propose a mutual guidance network for image enhancement. 
%As illustrated in the first rows in Fig.~\ref{fig:framework}, it consists of two components: The global branch, and the local branch. 
As illustrated in Figure~\ref{fig:framework}, the proposed network starts with a transformer- and CNN-head for each branch, respectively. 
%At first of the two branches, we use a Vit- and CNN-head for global and local feature extraction. 
This is for global and local feature extraction.
Then, a mutual guidance structure is developed for global-local fusion.

\noindent{\bf Global Feature extraction.} To enable the global image information reasoning while maintaining a shallow network architecture, we exploit a lightweight transformer (Vit) to regress a global feature~${\bf f}_g \in R^{C}$ with a given low-quality image $x$. The conventional transformers directly use Patch Embedding to obtain image tokens. However, with fixed patch sizes, it suffers from the quadratic increment of the image tokens. Instead, we first utilize an adaptive pooling operation applied on the high-resolution image $x \in R^{H\times W}$ to obtain a down-sized image $\hat{x} \in R^{S\times S}$. In this way, only a constant number of tokens are processed, which largely reduces the computational cost. After that, we use the multi-head attention to obtain long-range dependencies among those tokens. Finally, we transform the aggregated contents into a 1D global feature ${\bf f}_g$ using a single fully connected layer. We can formulate the above procedures as below:
\begin{align}
	\hat{x} &= {\rm AdaPooling}(x) \\
	{\bf f}_g &= {\rm Trans}(\hat{x})
\end{align}
With the introduced adaptive pooling, our transformer extraction module is able to handle images of arbitrary spatial resolutions at a constant computational cost.

\noindent{\bf Local Feature extraction.}
Unlike the global feature extraction, we obtain local image features from the original image $x$ at its full resolution, i.e.:
\begin{equation}
	{\bf F}_l = {\rm CNN}(x)
\end{equation}
where ${\bf F}_l$ represents local pixel-level features.

\noindent{\bf Mutual Guided Aggregation.}
In previous literature, global information represented as a 1D feature is directly fed into the accompanying local enhancement module for global-local fusion. It is inflexible and may lead to inferior performance since such one-shot fusion happening in the local feature extraction provides limited informative content for global-local adjustment.
In this paper, we present a novel mutual-guidance enhancement module. As shown in Figure~\ref{fig:framework}, apart from the CNN head and transformer head, we use $T(=5)$ parallel global-local interaction blocks for deep image feature extraction. At the end of each global-local interaction block pair, we perform a mutual guided aggregation for bidirectional information exchange. 

Given the global image feature ${\bf f}_g^{t}\in R^{C}$ and local image feature ${\bf F}_l^{t} \in R^{C \times h \times w }$ extracted by the $t$-th global-local interaction blocks, we conduct mutual guided aggregation between global and local contents. In the existing works, the conventional self-attention or sliced window attention will lead to ${\mathcal O}(w \times h \times C)$ computational complexity. To reduce the computation overhead, we propose a novel global-local attention mechanism. To begin with, we adopt a global pooling to obtain a 1D local feature ${\bf f}_l^{t} \in R^{C}$ from ${\bf F}_l^{t}$:
\begin{equation}
	{\bf f}_l^{t}= {\rm GlobalPooling}({\bf F}_l^{t}). \\
\end{equation}
After that, we perform bidirectional global-local attention operations~(${\rm Atten}(\cdot)$):
\begin{equation}
	\begin{split}
		&	w_{g \rightarrow l}^{t} = {\rm Atten}({\bf f}_l^{t},{\bf f}_g^{t}), \\
		&	w_{l \rightarrow g}^{t} = {\rm Atten}({\bf f}_g^{t},{\bf f}_l^{t}), \\
	\end{split}
	\label{atten}
\end{equation}
where $w_{f \rightarrow g}^{t}$ and $w_{l \rightarrow g}^{t}$ is the corresponding similarity scores between the current global and local features ${\bf f}_g^{t},{\bf f}_l^{t}$. 

In the view of computational efficiency, we propose a channel-wise attention mechanism for ${\rm Atten}({\bf f}_i,{\bf f}_j)$: 
\begin{equation}
	\begin{split}
		&	{\bf q}_i,{\bf k}_j,{\bf v}_j ={\rm FC}({\bf f}_i), {\rm FC}({\bf f}_j),{\rm FC}({\bf f}_j), \\
		&   {\rm Atten}({\bf f}_i,{\bf f}_j) = {\rm softmax}(\frac{{\bf q}_i {\bf k}_j^T}{{\rm \sqrt{d}}} ) {\bf v}_j,
	\end{split}
	\label{atten1}
\end{equation}
where ${\rm \sqrt{d}}$ is a scaling factor for normalization. ``${\rm FC}$" denotes a fully connected layer.
At last, we update the global and local image features:
\begin{equation}
	\begin{split}
		&	{\bf \hat F}_l^{t} = w_{g \rightarrow l}^{t} \odot {\bf F}_l^{t},\\
		&	{\bf \hat f}_g^{t} = w_{l \rightarrow g}^{t} {\bf f}_g^{t}, \\
	\end{split}
\end{equation}
where $\odot$ means channel-wise multiplication.
\\
The global and local results ${\bf \hat f}_g^{t}, {\bf \hat F}_l^{t}$ are taken as inputs for the next global-local interaction block for further processing.
By stacking multiple~($T$) mutual guidance interaction blocks, it generates the final global and local features ${\bf \hat f}_g^{T},{\bf \hat F}_l^{T}$.
%With our mutual-attention mechanism, not only does the global contextual information guides the local enhancement but also the local feature helps to refine the global contents.
With our mutual-attention mechanism, global contextual information can guide the local enhancement. In return, local feature also helps to refine the global contents.
Consequently, our global and local features are effectively fused, ensuring detailed enhancement with global consistency. 

\noindent{\bf Discussion}
Our design strikes a sweet point between performance and efficiency. In detail, we iteratively perform the fusion between $N$ global and $N$ local tokens ($N$ is pre-defined), regardless of the image resolutions. It largely reduces the computational cost for high-resolution images, while allowing for balanced global-local interactions.
There are some key differences between Mao~\etal~\cite{mao2021dual} and ours. (1) They conduct computationally intensive interaction between $H\times W$ and $N$ tokens for high-resolution cases. (2) Considering the number of local tokens ($H \times W$) is significantly larger than the global one ($N$), the local branch may dominate the fusion process, making mutual fusion less effective. (3) This strategy will introduce inconsistency of token numbers between training and testing, raising the risk of performance drop. Conformer~\cite{peng2021local} also explores a dual-stream model realized by convolutional layers. However, as discussed in PAC~\cite{su2019pixel}, the spatially invariant nature of convolutions may not fully leverage mutual information.

\noindent{\bf Residual Regression.}
We take the fused feature ${\bf F}_l^{T}$ as input and utilize a single convolution for residual regression:
\begin{equation}
	R_x = {\rm Conv}({\bf F}_l^{T}),
\end{equation}
$R_x$ is the output of our MGN that generally represents the difference between the source input and its target.

\subsection{Coarse-to-fine Residual Integration (C2F-RI)} \label{IRR}
\begin{figure}[t]
	\centering
	
	\includegraphics[width=1.0\linewidth]{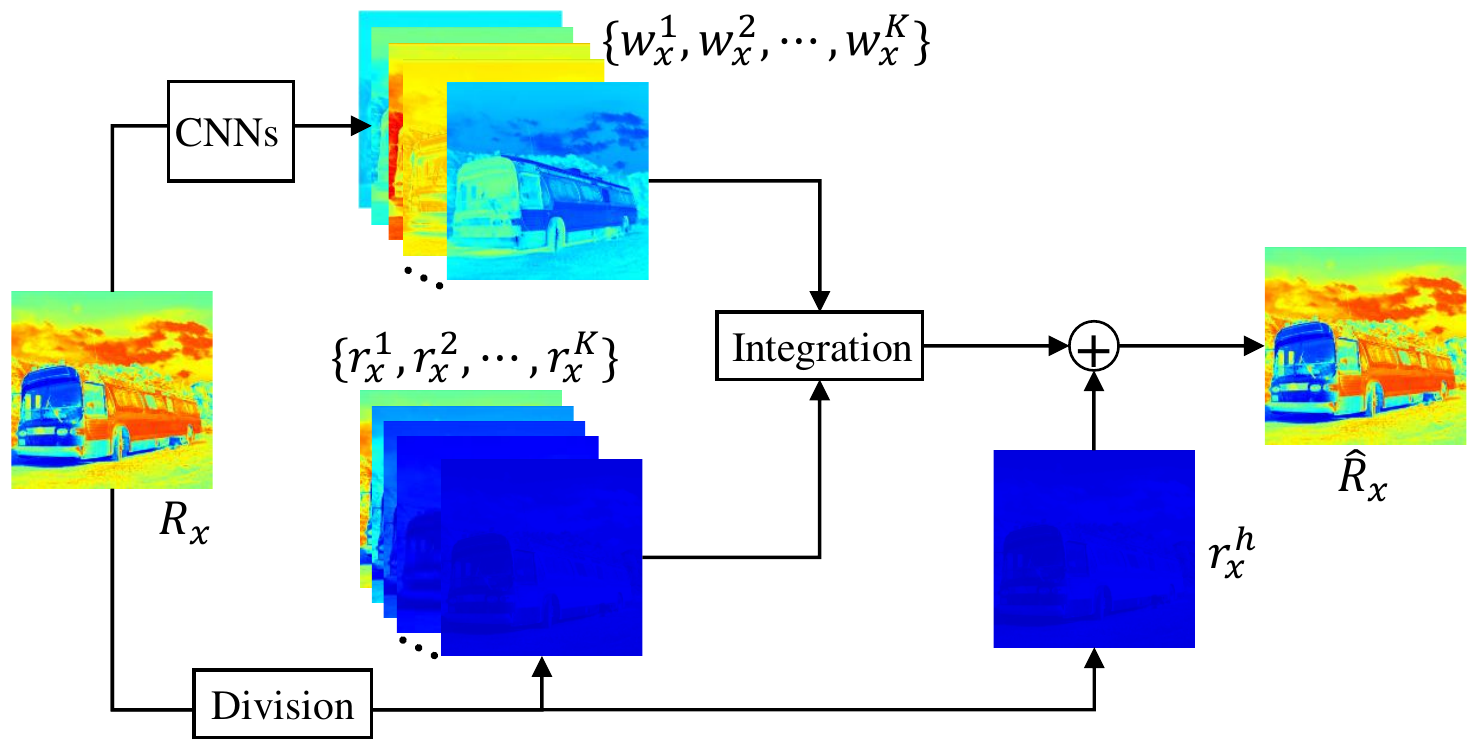} % 
	\caption{The pipeline of our C2F-RI. We first divide the $R_x$ into several residual pieces $\{r_x^{1},r_x^{2},\cdots,r_x^{k},\cdots r_x^{K}\}$ coarse-to-finely. Then, for each of the residual pieces $r_x^{k}$, we calculate the weighing map $w_x^k$ and perform integration to obtain a refined enhancement residue~${\hat R}_x$.  }
	\label{fig:highorder}
\end{figure} % 
%---------------------------------------------------------------------------------------------------------------
Residues produced by deep neural networks carry complex informative contents that help to fill the gap between source and target domains. Previous methods adopt the widely used residual learning that adds the learned residues with network inputs. However, the learning of image enhancement is highly non-linear, which is difficult to optimize directly. Inspired by human experts that usually carry out coarse-to-fine adjustments, we divide the residue into several pieces and predict importance maps for adaptive integration.
As demonstrated in Figure~\ref{fig:highorder}, there are two primary steps: residual division and integration.

%	We observe that the magnitude of adjustment required for a low-quality image actually varies severely in different image regions, making it difficult to restore the local details of a low-quality image in a single-shot aggregation manner. Thus, we propose a residual integration scheme built upon the divide-and-conquer strategy. As demonstrated in Figure~\ref{fig:highorder}, there are two main steps: residual division and integration.

\noindent{\bf Residual Division.}
Given a learned overall residue $R_x$ from our mutual guidance network, we first coarse-to-finely divide it into several~($K$) residual pieces $\{r_x^{1},r_x^{2},\cdots,r_x^{k},\cdots, r_x^{K}\}$ and a high-order rest term $r_x^{h}$:
\begin{equation}
	\begin{split}
		&	r_x^{k} = \frac{R_x}{2^k}, \text{and,  } r_x^{h} = \frac{R_x}{2^K}, \\
		&	R_x = \sum_{k=1}^{K} r_x^{k} + r_x^{h}
	\end{split}
	\label{residualpart}
\end{equation}

\noindent{\bf Residual Integration.}
Then, our residual integration is formulated as:
\begin{equation}
	y =  \sum_{k=1}^{K} r_x^{k} w_x^k + r_x^{h} + x,
\end{equation}
where $w_x^k$ is the $k$-th weighting map with respect to $r_x^{k}$, and $y$ is the enhanced result of our model. Note that, the directly reside learning is a special case of the above formulation, where $R_x$ is evenly divided $r_x^{k} = \frac{R_x}{K}$ with all weighting elements equal to 1. Therefore, our C2F-RI is more general. Note that, there may
be other models for such a residual partition, and we only utilize this route to evaluate and justify the feasibility of our idea.

\begin{table*}[ht]
	\renewcommand\arraystretch{1.1}
	\begin{center}
		\resizebox{\textwidth}{!}{
			\begin{tabular}{ c | c | c c c  c  c  c  c  c  c }
				%\begin{tabular}{ c | c | c | c | c | c | c | c | c | c | c }
				\hline
				Dataset & Metric 	 	&U-Net	 &HDRNet	 &DPE  	&DPED	 &DeepUPE &DeepLPF &CURL	&Star & \textbf{Ours} \\
				
				\hline
				
				%& LPIPS$\downarrow$ &  0.3026 & 0.2917 & 0.2903 & 0.3129 & 0.2755 & 0.2377 & 0.2529 & 0.2526 & \textcolor{red}{0.2202} \\
				%	~ & Params.$\downarrow$ 		&1.3M	 &-			&3.3M	&-		& 1.0M	& 800K	& 1.4M	&{\bf 27K}	& 411K		 \\
				MIT-Adobe-5K & PSNR$\uparrow$ 		&21.57	 &21.96		&22.15	&21.76		 &23.04 &24.48   &24.20 &24.46 & {\bf25.59}	 \\
				UPE & SSIM$\uparrow$		&0.843	 &0.866 	&0.850	&0.871		 &0.893	&0.887 &0.880	&0.919		 &{\bf0.930 }	 \\
				\hline \hline
				
				Dataset & Metric	 	&FCN	 &CRN	&U-NET &DPE  	&DPED	  &DeepLPF &CURL	&Star  & Ours	\\ \hline 
				
				%& LPIPS$\downarrow$ &  0.3026 & 0.2917 & 0.2903 & 0.3129 & 0.2755 & 0.2377 & 0.2529 & 0.2526 & \textcolor{red}{0.2202} \\
				%	~ & Params.$\downarrow$ 		&-	 &-		& -	&3.3M	&-		& 1.0M	& 800K 	&{\bf 27K}	& {\bf 411K}		 \\
				MIT-Adobe-5K & PSNR$\uparrow$ 		&20.66	 &22.38	&22.13	&23.80	&21.79		  &23.93   &24.04 &23.11 & {\bf24.53}		 \\
				DPE & SSIM$\uparrow$ 		&0.849	 &0.877 	&0.879	&0.900	&0.871		 	&0.903	&0.900	&0.903 &{\bf0.924 }	 \\
				\hline \hline
				
				Dataset & Metric	&3DLUT	 	 	&DeepUPE  	&DeepLPF	 &EG 	&MIR-Net 	&Uformer	&Star &Restormer  & Ours	\\ \hline 
				
				%& LPIPS$\downarrow$ &  0.3026 & 0.2917 & 0.2903 & 0.3129 & 0.2755 & 0.2377 & 0.2529 & 0.2526 & \textcolor{red}{0.2202} \\
				%	~ & Params.$\downarrow$ &-		&-	 			&-		&1.0M		&800K		& -		& 31.8M		& 38.9M	& {\bf 411K} \\
				LOL-v2 & PSNR$\uparrow$ 	&18.04		 		&15.08		&16.02		 &16.57 &21.94   	&19.66 &20.97 &22.42 & {\bf23.78}	 \\
				Synthetic & SSIM$\uparrow$ &0.800			 	&0.623		&0.578		 &0.734	&0.876		&0.871	&0.870 &0.861 &{\bf0.909 }	 \\
				\hline \hline
				
				Dataset & Metric			&3DLUT	 	&SRIE 	&DeepUPE  		  &IPT	&MIR-Net   &Uformer	&Star	&Restormer  	& Ours	\\ \hline

				%	~ & Params.$\downarrow$ &	-		&-	 		& -		&1.0M			& 31.8M	&38.9M	&-	&26.13	& {\bf 411K}		 \\
				LOL-v2 & PSNR$\uparrow$ 		&17.59	 		&17.34	&13.27	&19.80	&20.02  &18.82 	&18.26  &21.72	 &{\bf22.57}	 \\
				Real & SSIM$\uparrow$ 			&0.721	 	&0.686	&0.452		&0.813  &0.820		&0.771	&0.546 &{\bf0.823} &0.788 		\\
				\hline
				%\hline
		\end{tabular}}
	\end{center}
	
	\caption{Quantitative comparison of SOTAs on benchmarks.  `$\uparrow$' indicates the higher, the better. The best results are in {\bf bold}.}
	\label{tab:quan}
\end{table*}

\noindent{\bf Analysis.} In this part, motivated by the idea of divide-and-conquer that decomposes a complex problem into several easier sub-problems, we propose a residual integration scheme to exploit the informative residues produced by our mutual guidance network. To this end, we solely utilize a unified one-shot framework, while achieving coarse-to-fine image enhancement. Due to the flexibility and effectiveness of our piece-wise integration, it produces pleasing results with better global consistency.

\subsection{Total Loss}\label{loss}
\noindent{\bf Auxiliary Global-Local Supervisions.}
In our mutual guidance enhancement, we explore an effective global-local aggregation method for image enhancement. To enforce explicitly global and local learning, we introduce intermediate supervisions~(as shown in Figure~\ref{fig:framework}). Accordingly, we use an additional global regression block implemented by a fully connected layer to produce the global enhanced image~$x_g$:
\begin{equation}
	x_g = {\rm exp}(x + \epsilon,{\rm FC}({\bf f}_g^{T})) ,
\end{equation}
where ${\bf f}_g^{T}$ is the output of the global branch and $\epsilon$ is a small constant~($1e^{-6}$). 
Meanwhile, we employ the straightforward residual learning to obtain a local enhanced result: $x_l = x + R_x$.
Accordingly, apart from the final loss~$L_{f}$ between  $y$ and $y_{GT}$,  we introduced two additional loss terms to jointly optimize our networks. Our total loss can be expressed as follows:
\begin{equation}
	% ||y-y_{GT}||_{1}
	\begin{split}
		L_{total} & =L_{f}  + L_{aux},  L_{f} = ||y-y_{GT}||_{1},\\ 
		L_{aux}& = \alpha_g  L_g + \alpha_l  L_l , \\
		& = \alpha_g  ||x_g-y_{GT}||_{1} +  \alpha_l ||x_l-y_{GT}||_{1}. \\
	\end{split}
	\label{fho}
\end{equation}
$y_{GT}$ is the ground-truth image with respect to the input image $x$.  We set $\alpha_g=0.01$ and $ \alpha_l=0.05$ through a grid hyper-parameter searching. $||\cdot||_1$ refers to the L1 loss.

\section{Experiments}\label{sec:exp}

\subsubsection{Implementation Details.}
All of our models are trained from scratch with a batchsize of 8. The input size is $128 \times 128$. Vertical or horizontal flipping, and $90^{\circ}$ rotation are randomly applied for data augmentation. An Adam optimizer is adopted and the learning rate decays from $5 \times 10^{-4}$ to $0$ by a cosine annealing strategy. The whole training phase lasts for 300 epochs, each of which contains 1000 iterations. The training time is about 6 hours with a single NVIDIA RTX 2080 Ti GPU. The detailed network structure is illustrated in our supplementary materials.

\subsection{Datasets and Metrics}

% %------------------------------------------------------------------------

\subsubsection{MIT-Adobe-5K-UPE}
We process the MIT-Adobe-5K-UPE dataset as described in~\cite{Wang_2019_CVPR}. In total, it consists of 5000 high-quality RGB image pairs. We use the most common protocols like~\cite{Wang_2019_CVPR,chen2018deep,moran2020deeplpf,moran2021curl}, and split the whole image pairs into three parts: 4000 samples to train our model, another 500 images for evaluation and the remaining part is used as the testing set for quantitative comparison. The retouched images by expert C are used as GTs.

\subsubsection{MIT-Adobe-5K-DPE}
It contains 5000 images captured by DSLR cameras and manually adjusted by five experts. Following the previous methods~\cite{moran2020deeplpf,moran2021curl,song2021starenhancer,chen2017photographic,chen2017fast,ignatov2017dslr}, we take the images enhanced by Artist C as the target and use the 2250 image pairs for training, 2250 image pairs for validation, and the remaining 500 images for testing.

\subsubsection{LOL-v2 Synthetic}
It is a noise-less synthetic dataset~\cite{yang2021sparse}, in which 900 low/normal light image pairs are used for training and another 100 images for evaluation.

\subsubsection{LOL-v2 Real}

LOL-v2 Real~\cite{yang2021sparse} is a set of captured low/normal light images which have severe noise. Following previous methods~\cite{yang2020fidelity,Zamir2020MIRNet,yang2021sparse}, we utilize the split parts composed of 689 image pairs for training and the remaining 100 samples for testing.

\subsubsection{Metrics}
Peak Single-to-Noise Ratio~(PSNR) and Structural Similarity Index (SSIM) are employed for evaluating the quality of the enhanced images. 
%---------------------------------------------------------------------------------------------------------------

\begin{figure*}[t]
	\centering
	
	\includegraphics[width=1.0\linewidth]{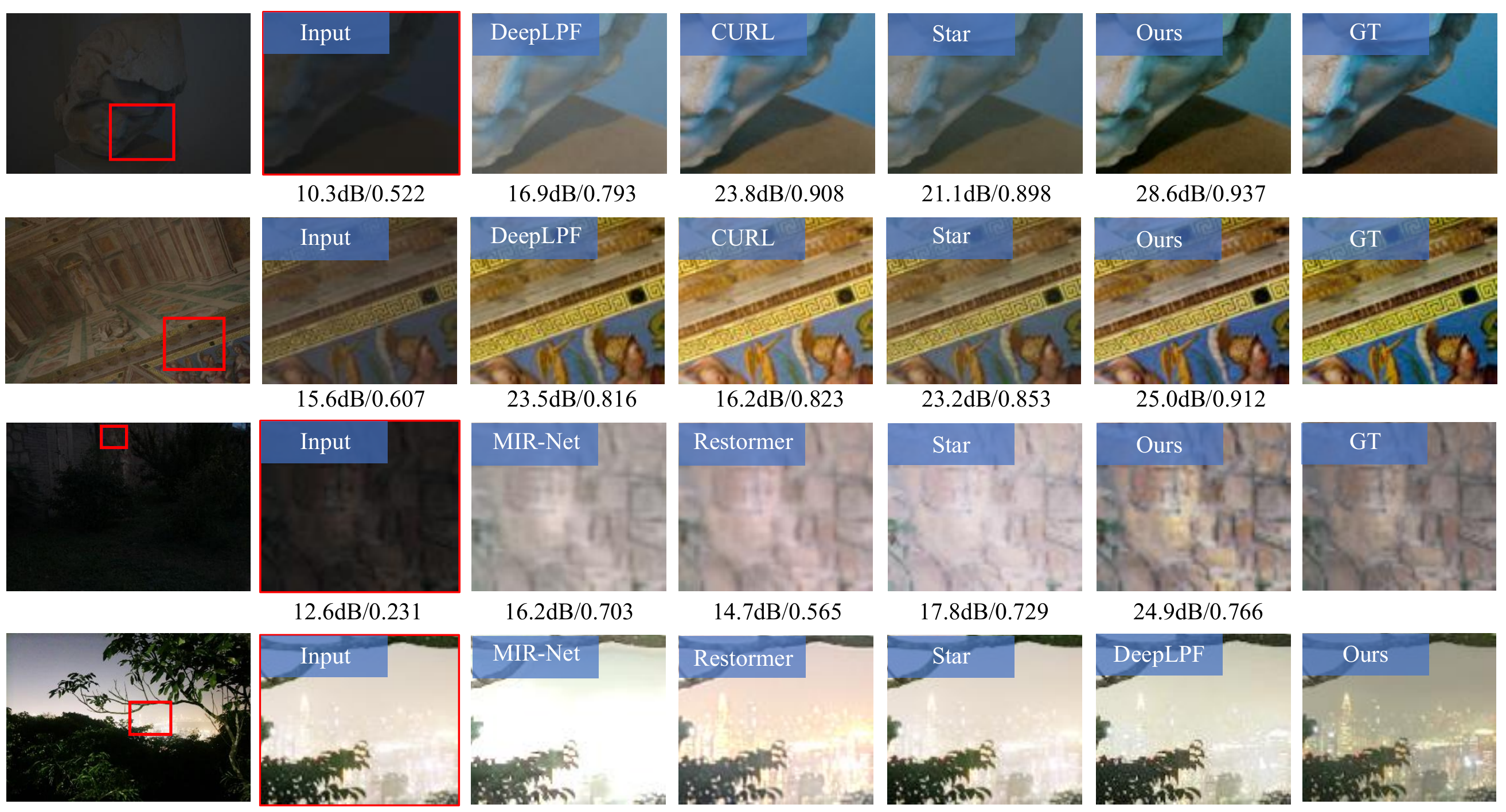} 
	\caption{Qualitative Comparison on Adobe-5K-UPE~\protect\cite{Wang_2019_CVPR}, MIT-Adobe-5K-DPE~\protect\cite{chen2018deep}, LOL-v2 Real~\protect\cite{yang2021sparse} and a real-world case captured by a mobile phone. In the first three rows, the values beneath images represent the PSNR~(dB) and SSIM. } 
	\label{fig:sota_viz}
\end{figure*} % 

\subsection{Comparison with State-of-the-Art Approaches}

We make quantitative and qualitative comparison with competitive SOTAs, including Deep-UPE~\cite{Wang_2019_CVPR}, CURL~\cite{moran2021curl}, DeepLPF~\cite{moran2020deeplpf}, MIR-Net~\cite{Zamir2020MIRNet}, 3DLUT~\cite{zeng2020learning}, IPT~\cite{chen2021pre}, Uformer~\cite{wang2021uformer}, Star~\cite{zhang2021star} and Restormer~\cite{zamir2022restormer}. Following these previous works, we use MIT-Adobe-5K-UPE~\cite{Wang_2019_CVPR}, MIT-Adobe-5K-DPE~\cite{chen2018deep}, LOL-v2 Synthetic and LOL-v2 Real~\cite{yang2021sparse} for quantitative evaluation while including real-world cases for qualitative analysis.

\subsubsection{Quantitative Comparison.}
Table~\ref{tab:quan} reports the results of image enhancement methods on representative benchmarks. It shows that our method yields new state-of-the-art in terms of PSNR. Specifically, our model outperforms these transformer-based models~(Uformer~\cite{wang2021uformer}, IPT~\cite{chen2021pre}, Restormer~\cite{zamir2022restormer}, Star~\cite{zhang2021star}) as well as the large CNN-based methods~(Deep-UPE~\cite{Wang_2019_CVPR},  MIR-Net~\cite{Zamir2020MIRNet}). 
Since some large models with millions of parameters like~{MIR-Net (31.8M), IPT (115M) and Restormer (26.13M)}, that are 64-287 times larger than our model~(0.4M), are capable of removing serve image noise, they get higher SSIM values than those lightweight models~(3DLUT, Star and Ours) in LOL-v2 Real. Nevertheless, their PSNRs are consistently lower than ours in all datasets by a large margin. 
% Although MIR-Net, DRBN and Restormer has higher SSIM in dataset LOL Real, their model size is xxx times larger than our model.
The experiments demonstrate the effectiveness and efficiency of the proposed algorithm.

\subsubsection{Qualitative Comparison}
In addition, we present some visual examples in Figure~\ref{fig:sota_viz}. The competing SOTA methods fail to predict natural image colors or appropriately brighten dark areas, yielding degraded image qualities. In contrast, with the proposed mutual guidance aggregation scheme, our method is capable of recovering pleasing image contents in the view of local contrast and global illumination condition. We also conduct a user study and give more visual examples for evaluation in the supplementary materials.
\begin{figure}[t]
	\centering
	\includegraphics[width=1.0\linewidth]{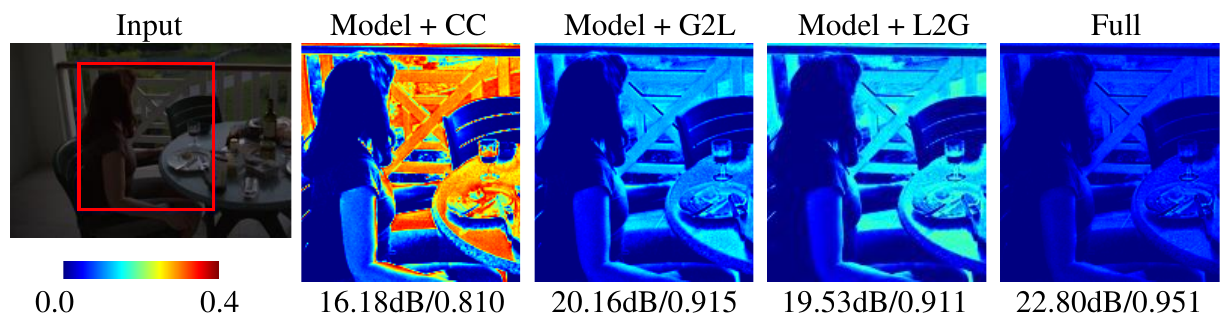} % 
	\caption{Analysis of the mutual guidance attention. We draw the normalized L2 distances between predictions and the GT for comparison. The numbers are PSNR and SSIM.}
	\label{fig:mutualviz} 
\end{figure} %
\begin{table}[t]
	\begin{center}
		\begin{tabular}{l|c|c|c|c } %p{2cm}p{2cm}
			\hline
			Models~(Param.)		&CC         &L2G         &G2L     &PSNR/SSIM        \\ \hline
			
			Model + CC~(413K)	&\cmark     &       	&           & 23.63/0.902              \\       
			Model + L2G~(411K)	&	        & \cmark    &           & 24.98/0.923              \\      
			Model + G2L~(411K)	&      		&       	&\cmark           & 24.85/0.921            \\  
			Full~(411K)		&	        & \cmark    &\cmark           & {\bf 25.59}/{\bf0.930}            \\  \hline    
		\end{tabular}
	\end{center}
	\caption{Impacts of our mutual guidance structure for image enhancement. ``CC" refers to adopting channel concatenation for global-local information fusion. ``G2L" and ``L2G" means employing either global-to-local or local-to-global guidance. The full model is with the proposed mutual guidance modeling. The model complexities are also reported.}
	\label{table:ablation}
\end{table}
\subsection{Ablation Study}
To better understand how each design affects the final performance, we conduct extensive ablation studies on MIT-Adobe-5K-UPE. More detailed ablating analyses are presented in our supplementary materials.

\subsubsection{Impact of Mutual Guidance.} 
Here, we conduct experiments to evaluate the contribution of our mutual guidance modeling. Keeping other parts of our framework unchanged, we start from the baseline that solely replaces the mutual guidance aggregation with the widely used channel concatenation strategy~(named ``Model + CC"). Then, we train two additional models with either global-to-local or local-to-global fusion~(dubbed as ``Model + G2L/L2G"). Finally, our full model utilizes the proposed mutual guidance aggregation. As shown in Table~\ref{table:ablation}, both the ``Model + G2L" and ``Model + L2G" surpass the ``Model + CC" and achieve 1.35dB/1.22dB improvement in terms of PSNR, indicating the importance of global-local information exchange. Especially, our full model with mutual guidance aggregation achieves significant improvement. Apart from the ``Model + CC", the other three models keep an identical framework~(but different fusion strategies), yielding the same model parameters. Therefore, it concludes that the performance gains are mainly attributed to our mutual guidance modeling. Moreover, Figure~\ref{fig:mutualviz} shows our full model achieves the smallest L2 error. All these results manifest the effectiveness and efficiency of our mutual guidance structure.

\begin{figure}[t]
	\centering
	\includegraphics[width=1.0\linewidth]{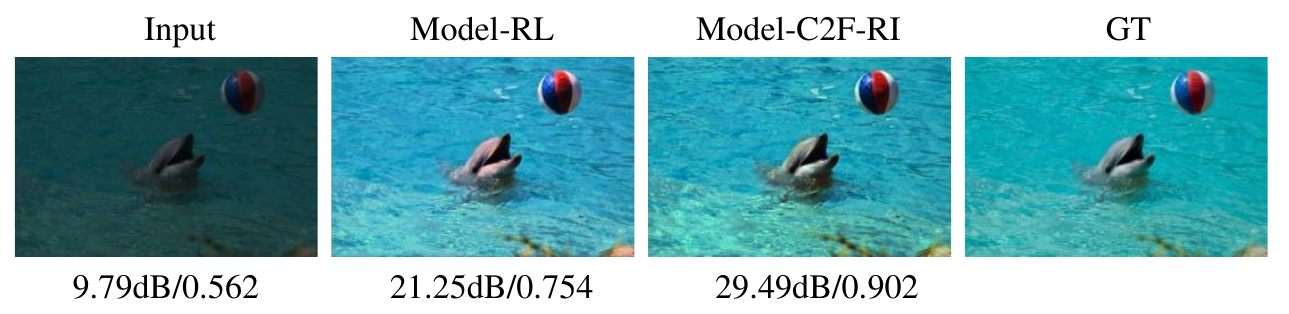} % 
	\caption{Qualitative comparison between the residual learning and the proposed C2F-RI. }
	\label{fig:rl_c2fri} 
	% mutual_ablation.png
\end{figure} % 

\begin{table}[h]
	\begin{center}
		\begin{tabular}{l|c|c|c } %p{2cm}p{2cm}
			\hline
			Models		&PSNR         &SSIM         &Param.        \\ \hline
			Model-RL 		&24.94         &0.912         &405K           \\ \hline
			Model-C2F-RI		&{\bf 25.59}         &{\bf 0.930}            &411K    \\ \hline
			
		\end{tabular}
	\end{center}
	\caption{Ablation study of our C2F-RI.}
	\label{table:rlablation}
\end{table}

\begin{table}[h]
	\begin{center}
		\begin{tabular}{l|c|c|c|c|c|c } %p{2cm}p{2cm}
			\hline
			$K$		&1         &2         &4     &8 	&12 	&16        \\ \hline
			PSNR 		&23.50         &24.95         &25.38     &{\bf 25.59} 	&25.24 	&25.29        \\ \hline
			SSIM		&0.881         &0.916         &0.927     &{\bf 0.930} 	&0.925 	&0.926        \\ \hline
			
		\end{tabular}
	\end{center}
	\caption{Impact of different partition numbers~($K$) for residual division in our C2F-RI.}
	\label{table:partition}
\end{table}

\subsubsection{Residual Learning vs. C2F-RI.} To understand the influence of our coarse-to-fine residual integration design, we ablate our C2F-RI and adopt widely used residual learning for image enhancement~(named ``Model-RL"). From Table~\ref{table:rlablation}, it can be clearly seen that Model-C2F-RI outperforms Model-RL with 0.65dB/0.018 improvements in terms of PSNR and SSIM. Figure~\ref{fig:rl_c2fri} shows our Model-C2F-RI generates better results than Model-RL~(see the dolphin).

\subsubsection{Partition Number in C2F-RI.}
In default, we adopt Eq.~\ref{residualpart} to split $R_x$ into 8 residual parts. Here, we explore the connection between the final performance and the partition number. We train six models adopting different partition numbers $K = \{1,2,4,8,12,16\}$. Especially, $K=1$ means no division step is performed. According to the results in Table~\ref{table:partition}, compared with this model without residual partition, the other models~($K>1$) taking the advantage of the C2F-RI strategy obtain superior performance. This comparison verifies the effectiveness of our C2F-RI. Meanwhile, we observe that the performance is not consistently positively-correlated with the partition number. With the increment of $K$, the information of high-order residue slices drops abruptly. As a result, it is difficult for the network to precisely predict associated importance maps for high-order content aggregation. In contrast, the default setting obtains the best result.

\begin{figure}[t]
	\centering
	
	\includegraphics[width=1.0\columnwidth]{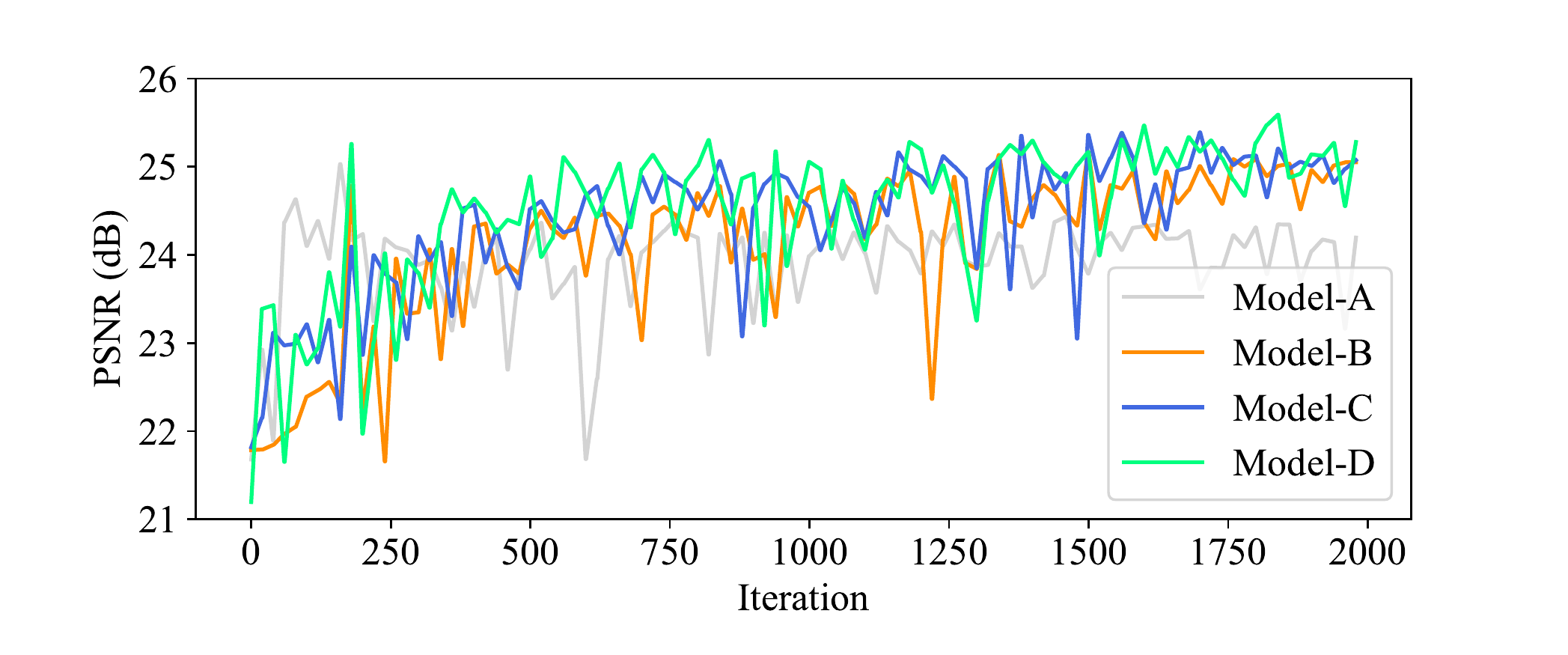} 
	\caption{The validation PSNR curves of using different training losses. ``Model-A" refers to the model trained with only the final loss~$L_{f}$. ``Model-B/C" means using global/local auxiliary supervision, respectively. ``Model-D" is our full supervision that is trained with all of the three losses. } 
	\label{fig:loss_curves}
	
\end{figure}

\subsubsection{The Auxiliary Global-Local Loss}
As described in our method section, we include two additional loss terms for intermediate supervision. We expect such supervision could help for explicitly global and local contextual learning. To verify our design, we train several variants that employ different combinations of the three loss terms and evaluate these models quantitatively. The validation curves~in Figure~\ref {fig:loss_curves} illustrate that both the global and local auxiliary terms are able to boost the final performance, indicating the necessity of auxiliary global-local supervisions.

%------------------------------------------------------------------------
\section{Conclusion}
In this paper, we propose mutual guidance modeling and an integral residual learning method for image enhancement. Our model realizes an effective global-local information fusion through stacked mutual guidance attention blocks.  Taking the advantage of the proposed residual integration, it is capable of precisely refining the predicted residues, achieving better performance and pleasing visual enhancement results. Extensive quantitative and qualitative experiments manifest the effectiveness of our proposed method. 
.

%% The file named.bst is a bibliography style file for BibTeX 0.99c
\bibliographystyle{named}
\bibliography{ARXIV-MGNLLE}

\begin{thebibliography}{}

\bibitem[\protect\citeauthoryear{Chen and Koltun}{2017}]{chen2017photographic}
Qifeng Chen and Vladlen Koltun.
\newblock Photographic image synthesis with cascaded refinement networks.
\newblock In {\em Proceedings of the IEEE international conference on computer
  vision}, pages 1511--1520, 2017.

\bibitem[\protect\citeauthoryear{Chen \bgroup \em et al.\egroup
  }{2017}]{chen2017fast}
Qifeng Chen, Jia Xu, and Vladlen Koltun.
\newblock Fast image processing with fully-convolutional networks.
\newblock In {\em Proceedings of the IEEE International Conference on Computer
  Vision}, pages 2497--2506, 2017.

\bibitem[\protect\citeauthoryear{Chen \bgroup \em et al.\egroup
  }{2018}]{chen2018deep}
Yu-Sheng Chen, Yu-Ching Wang, Man-Hsin Kao, and Yung-Yu Chuang.
\newblock Deep photo enhancer: Unpaired learning for image enhancement from
  photographs with gans.
\newblock In {\em Proceedings of the IEEE Conference on Computer Vision and
  Pattern Recognition}, pages 6306--6314, 2018.

\bibitem[\protect\citeauthoryear{Chen \bgroup \em et al.\egroup
  }{2021}]{chen2021pre}
Hanting Chen, Yunhe Wang, Tianyu Guo, Chang Xu, Yiping Deng, Zhenhua Liu, Siwei
  Ma, Chunjing Xu, Chao Xu, and Wen Gao.
\newblock Pre-trained image processing transformer.
\newblock In {\em CVPR}, pages 12299--12310, 2021.

\bibitem[\protect\citeauthoryear{Cui \bgroup \em et al.\egroup
  }{2022}]{cui2022illumination}
Ziteng Cui, Kunchang Li, Lin Gu, Shenghan Su, Peng Gao, Zhengkai Jiang,
  Yu~Qiao, and Tatsuya Harada.
\newblock Illumination adaptive transformer.
\newblock {\em arXiv preprint arXiv:2205.14871}, 2022.

\bibitem[\protect\citeauthoryear{Dosovitskiy \bgroup \em et al.\egroup
  }{2020}]{dosovitskiy2020image}
Alexey Dosovitskiy, Lucas Beyer, Alexander Kolesnikov, Dirk Weissenborn,
  Xiaohua Zhai, Thomas Unterthiner, Mostafa Dehghani, Matthias Minderer, Georg
  Heigold, Sylvain Gelly, et~al.
\newblock An image is worth 16x16 words: Transformers for image recognition at
  scale.
\newblock {\em arXiv preprint arXiv:2010.11929}, 2020.

\bibitem[\protect\citeauthoryear{Gharbi \bgroup \em et al.\egroup
  }{2017}]{gharbi2017deep}
Micha{\"e}l Gharbi, Jiawen Chen, Jonathan~T Barron, Samuel~W Hasinoff, and
  Fr{\'e}do Durand.
\newblock Deep bilateral learning for real-time image enhancement.
\newblock {\em ACM Transactions on Graphics (TOG)}, 36(4):1--12, 2017.

\bibitem[\protect\citeauthoryear{Guan \bgroup \em et al.\egroup
  }{2009}]{guan2009image}
Xu~Guan, Su~Jian, Pan Hongda, Zhang Zhiguo, and Gong Haibin.
\newblock An image enhancement method based on gamma correction.
\newblock In {\em 2009 Second international symposium on computational
  intelligence and design}, volume~1, pages 60--63. IEEE, 2009.

\bibitem[\protect\citeauthoryear{Hu \bgroup \em et al.\egroup
  }{2018}]{hu2018exposure}
Yuanming Hu, Hao He, Chenxi Xu, Baoyuan Wang, and Stephen Lin.
\newblock Exposure: A white-box photo post-processing framework.
\newblock {\em ACM Transactions on Graphics (TOG)}, 37(2):1--17, 2018.

\bibitem[\protect\citeauthoryear{Ignatov \bgroup \em et al.\egroup
  }{2017}]{ignatov2017dslr}
Andrey Ignatov, Nikolay Kobyshev, Radu Timofte, Kenneth Vanhoey, and Luc
  Van~Gool.
\newblock Dslr-quality photos on mobile devices with deep convolutional
  networks.
\newblock In {\em Proceedings of the IEEE International Conference on Computer
  Vision}, pages 3277--3285, 2017.

\bibitem[\protect\citeauthoryear{Kim \bgroup \em et al.\egroup
  }{2020}]{kim2020global}
Han-Ul Kim, Young~Jun Koh, and Chang-Su Kim.
\newblock Global and local enhancement networks for paired and unpaired image
  enhancement.
\newblock In {\em European Conference on Computer Vision}, pages 339--354.
  Springer, 2020.

\bibitem[\protect\citeauthoryear{Kim \bgroup \em et al.\egroup
  }{2021}]{kim2021representative}
Hanul Kim, Su-Min Choi, Chang-Su Kim, and Yeong~Jun Koh.
\newblock Representative color transform for image enhancement.
\newblock In {\em Proceedings of the IEEE/CVF International Conference on
  Computer Vision}, pages 4459--4468, 2021.

\bibitem[\protect\citeauthoryear{Kinoshita and
  Kiya}{2019}]{kinoshita2019convolutional}
Yuma Kinoshita and Hitoshi Kiya.
\newblock Convolutional neural networks considering local and global features
  for image enhancement.
\newblock In {\em 2019 IEEE International Conference on Image Processing
  (ICIP)}, pages 2110--2114. IEEE, 2019.

\bibitem[\protect\citeauthoryear{Lanchantin \bgroup \em et al.\egroup
  }{2021}]{lanchantin2021general}
Jack Lanchantin, Tianlu Wang, Vicente Ordonez, and Yanjun Qi.
\newblock General multi-label image classification with transformers.
\newblock In {\em CVPR}, pages 16478--16488, 2021.

\bibitem[\protect\citeauthoryear{Li \bgroup \em et al.\egroup
  }{2021}]{li2021lifting}
Wenhao Li, Hong Liu, Runwei Ding, Mengyuan Liu, and Pichao Wang.
\newblock Lifting transformer for 3d human pose estimation in video.
\newblock {\em arXiv preprint arXiv:2103.14304}, 2021.

\bibitem[\protect\citeauthoryear{Lin \bgroup \em et al.\egroup
  }{2021}]{lin2021end}
Kevin Lin, Lijuan Wang, and Zicheng Liu.
\newblock End-to-end human pose and mesh reconstruction with transformers.
\newblock In {\em CVPR}, pages 1954--1963, 2021.

\bibitem[\protect\citeauthoryear{Mao \bgroup \em et al.\egroup
  }{2021}]{mao2021dual}
Mingyuan Mao, Renrui Zhang, Honghui Zheng, Teli Ma, Yan Peng, Errui Ding,
  Baochang Zhang, Shumin Han, et~al.
\newblock Dual-stream network for visual recognition.
\newblock {\em Advances in Neural Information Processing Systems},
  34:25346--25358, 2021.

\bibitem[\protect\citeauthoryear{Moran \bgroup \em et al.\egroup
  }{2020}]{moran2020deeplpf}
Sean Moran, Pierre Marza, Steven McDonagh, Sarah Parisot, and Gregory Slabaugh.
\newblock Deeplpf: Deep local parametric filters for image enhancement.
\newblock In {\em Proceedings of the IEEE/CVF Conference on Computer Vision and
  Pattern Recognition}, pages 12826--12835, 2020.

\bibitem[\protect\citeauthoryear{Moran \bgroup \em et al.\egroup
  }{2021}]{moran2021curl}
McDonagh~Steven Moran, Sean, Mao Slabaugh, Gregorym, Liu Ren, Ying Tai, and
  Xiaoming Liu.
\newblock Curl: Neural curve layers for global image enhancement.
\newblock In {\em 2020 25th International Conference on Pattern Recognition
  (ICPR)}, pages 9796--9803. IEEE, 2021.

\bibitem[\protect\citeauthoryear{Park \bgroup \em et al.\egroup
  }{2018}]{park2018distort}
Jongchan Park, Joon-Young Lee, Donggeun Yoo, and In~So Kweon.
\newblock Distort-and-recover: Color enhancement using deep reinforcement
  learning.
\newblock In {\em Proceedings of the IEEE Conference on computer vision and
  pattern recognition}, pages 5928--5936, 2018.

\bibitem[\protect\citeauthoryear{Peng \bgroup \em et al.\egroup
  }{2021a}]{peng2021u}
Lintao Peng, Chunli Zhu, and Liheng Bian.
\newblock U-shape transformer for underwater image enhancement.
\newblock {\em arXiv preprint arXiv:2111.11843}, 2021.

\bibitem[\protect\citeauthoryear{Peng \bgroup \em et al.\egroup
  }{2021b}]{peng2021local}
Zhiliang Peng, Wei Huang, Shanzhi Gu, Lingxi Xie, Yaowei Wang, Jianbin Jiao,
  and Qixiang~Ye Conformer.
\newblock Local features coupling global representations for visual
  recognition. in 2021 ieee.
\newblock In {\em CVF International Conference on Computer Vision, ICCV}, pages
  357--366, 2021.

\bibitem[\protect\citeauthoryear{Pizer \bgroup \em et al.\egroup
  }{1990}]{pizer1990contrast}
Stephen~M Pizer, R~Eugene Johnston, James~P Ericksen, Bonnie~C Yankaskas, and
  Keith~E Muller.
\newblock Contrast-limited adaptive histogram equalization: speed and
  effectiveness.
\newblock In {\em [1990] Proceedings of the First Conference on Visualization
  in Biomedical Computing}, pages 337--338. IEEE Computer Society, 1990.

\bibitem[\protect\citeauthoryear{Shen \bgroup \em et al.\egroup
  }{2017}]{shen2017msr}
Liang Shen, Zihan Yue, Fan Feng, Quan Chen, Shihao Liu, and Jie Ma.
\newblock Msr-net: Low-light image enhancement using deep convolutional
  network.
\newblock {\em arXiv preprint arXiv:1711.02488}, 2017.

\bibitem[\protect\citeauthoryear{Song \bgroup \em et al.\egroup
  }{2021}]{song2021starenhancer}
Yuda Song, Hui Qian, and Xin Du.
\newblock Starenhancer: Learning real-time and style-aware image enhancement.
\newblock In {\em Proceedings of the IEEE/CVF International Conference on
  Computer Vision}, pages 4126--4135, 2021.

\bibitem[\protect\citeauthoryear{Souibgui \bgroup \em et al.\egroup
  }{2022}]{souibgui2022docentr}
Mohamed~Ali Souibgui, Sanket Biswas, Sana~Khamekhem Jemni, Yousri Kessentini,
  Alicia Forn{\'e}s, Josep Llad{\'o}s, and Umapada Pal.
\newblock Docentr: An end-to-end document image enhancement transformer.
\newblock {\em arXiv preprint arXiv:2201.10252}, 2022.

\bibitem[\protect\citeauthoryear{Su \bgroup \em et al.\egroup
  }{2019}]{su2019pixel}
Hang Su, Varun Jampani, Deqing Sun, Orazio Gallo, Erik Learned-Miller, and Jan
  Kautz.
\newblock Pixel-adaptive convolutional neural networks.
\newblock In {\em Proceedings of the IEEE/CVF Conference on Computer Vision and
  Pattern Recognition}, pages 11166--11175, 2019.

\bibitem[\protect\citeauthoryear{Wang \bgroup \em et al.\egroup
  }{2019}]{Wang_2019_CVPR}
Ruixing Wang, Qing Zhang, Chi-Wing Fu, Xiaoyong Shen, Wei-Shi Zheng, and Jiaya
  Jia.
\newblock Underexposed photo enhancement using deep illumination estimation.
\newblock In {\em The IEEE Conference on Computer Vision and Pattern
  Recognition (CVPR)}, June 2019.

\bibitem[\protect\citeauthoryear{Wang \bgroup \em et al.\egroup
  }{2021a}]{wang2021max}
Huiyu Wang, Yukun Zhu, Hartwig Adam, Alan Yuille, and Liang-Chieh Chen.
\newblock Max-deeplab: End-to-end panoptic segmentation with mask transformers.
\newblock In {\em CVPR}, pages 5463--5474, 2021.

\bibitem[\protect\citeauthoryear{Wang \bgroup \em et al.\egroup
  }{2021b}]{wang2021end}
Yuqing Wang, Zhaoliang Xu, Xinlong Wang, Chunhua Shen, Baoshan Cheng, Hao Shen,
  and Huaxia Xia.
\newblock End-to-end video instance segmentation with transformers.
\newblock In {\em CVPR}, pages 8741--8750, 2021.

\bibitem[\protect\citeauthoryear{Wang \bgroup \em et al.\egroup
  }{2021c}]{wang2021uformer}
Zhendong Wang, Xiaodong Cun, Jianmin Bao, and Jianzhuang Liu.
\newblock Uformer: A general u-shaped transformer for image restoration.
\newblock {\em arXiv preprint arXiv:2106.03106}, 2021.

\bibitem[\protect\citeauthoryear{Wang \bgroup \em et al.\egroup
  }{2022}]{wang2022structural}
Cong Wang, Jinshan Pan, and Xiao-Ming Wu.
\newblock Structural prior guided generative adversarial transformers for
  low-light image enhancement.
\newblock {\em arXiv preprint arXiv:2207.07828}, 2022.

\bibitem[\protect\citeauthoryear{Xu \bgroup \em et al.\egroup
  }{2020}]{xu2020learning}
Ke~Xu, Xin Yang, Baocai Yin, and Rynson~WH Lau.
\newblock Learning to restore low-light images via
  decomposition-and-enhancement.
\newblock In {\em Proceedings of the IEEE/CVF Conference on Computer Vision and
  Pattern Recognition}, pages 2281--2290, 2020.

\bibitem[\protect\citeauthoryear{Xu \bgroup \em et al.\egroup
  }{2022}]{xu2022snr}
Xiaogang Xu, Ruixing Wang, Chi-Wing Fu, and Jiaya Jia.
\newblock Snr-aware low-light image enhancement.
\newblock In {\em Proceedings of the IEEE/CVF Conference on Computer Vision and
  Pattern Recognition}, pages 17714--17724, 2022.

\bibitem[\protect\citeauthoryear{Yang \bgroup \em et al.\egroup
  }{2016}]{yang2016enhancement}
Jie Yang, Xinwei Jiang, Chunhong Pan, and Cheng-Lin Liu.
\newblock Enhancement of low light level images with coupled dictionary
  learning.
\newblock In {\em 2016 23rd International Conference on Pattern Recognition
  (ICPR)}, pages 751--756. IEEE, 2016.

\bibitem[\protect\citeauthoryear{Yang \bgroup \em et al.\egroup
  }{2020}]{yang2020fidelity}
Wenhan Yang, Shiqi Wang, Yuming Fang, Yue Wang, and Jiaying Liu.
\newblock From fidelity to perceptual quality: A semi-supervised approach for
  low-light image enhancement.
\newblock In {\em Proceedings of the IEEE/CVF Conference on Computer Vision and
  Pattern Recognition}, pages 3063--3072, 2020.

\bibitem[\protect\citeauthoryear{Yang \bgroup \em et al.\egroup
  }{2021}]{yang2021sparse}
Wenhan Yang, Wenjing Wang, Haofeng Huang, Shiqi Wang, and Jiaying Liu.
\newblock Sparse gradient regularized deep retinex network for robust low-light
  image enhancement.
\newblock {\em IEEE Transactions on Image Processing}, 30:2072--2086, 2021.

\bibitem[\protect\citeauthoryear{Zamir \bgroup \em et al.\egroup
  }{2020}]{Zamir2020MIRNet}
Syed~Waqas Zamir, Aditya Arora, Salman Khan, Munawar Hayat, Fahad~Shahbaz Khan,
  Ming-Hsuan Yang, and Ling Shao.
\newblock Learning enriched features for real image restoration and
  enhancement.
\newblock In {\em ECCV}, 2020.

\bibitem[\protect\citeauthoryear{Zamir \bgroup \em et al.\egroup
  }{2022}]{zamir2022restormer}
Syed~Waqas Zamir, Aditya Arora, Salman Khan, Munawar Hayat, Fahad~Shahbaz Khan,
  and Ming-Hsuan Yang.
\newblock Restormer: Efficient transformer for high-resolution image
  restoration.
\newblock In {\em Proceedings of the IEEE/CVF Conference on Computer Vision and
  Pattern Recognition}, pages 5728--5739, 2022.

\bibitem[\protect\citeauthoryear{Zeng \bgroup \em et al.\egroup
  }{2020}]{zeng2020learning}
Hui Zeng, Jianrui Cai, Lida Li, Zisheng Cao, and Lei Zhang.
\newblock Learning image-adaptive 3d lookup tables for high performance photo
  enhancement in real-time.
\newblock {\em IEEE Transactions on Pattern Analysis and Machine Intelligence},
  2020.

\bibitem[\protect\citeauthoryear{Zhang \bgroup \em et al.\egroup
  }{2018}]{zhang2018rcan}
Yulun Zhang, Kunpeng Li, Kai Li, Lichen Wang, Bineng Zhong, and Yun Fu.
\newblock Image super-resolution using very deep residual channel attention
  networks.
\newblock In {\em ECCV}, 2018.

\bibitem[\protect\citeauthoryear{Zhang \bgroup \em et al.\egroup
  }{2021}]{zhang2021star}
Zhaoyang Zhang, Yitong Jiang, Jun Jiang, Xiaogang Wang, Ping Luo, and Jinwei
  Gu.
\newblock Star: A structure-aware lightweight transformer for real-time image
  enhancement.
\newblock In {\em Proceedings of the IEEE/CVF International Conference on
  Computer Vision}, pages 4106--4115, 2021.

\bibitem[\protect\citeauthoryear{Zheng \bgroup \em et al.\egroup
  }{2021}]{zheng2021rethinking}
Sixiao Zheng, Jiachen Lu, Hengshuang Zhao, Xiatian Zhu, Zekun Luo, Yabiao Wang,
  Yanwei Fu, Jianfeng Feng, Tao Xiang, Philip~HS Torr, et~al.
\newblock Rethinking semantic segmentation from a sequence-to-sequence
  perspective with transformers.
\newblock In {\em CVPR}, pages 6881--6890, 2021.

\end{thebibliography}

\renewcommand\thesection{\Alph{section}}
\renewcommand\thesubsection{\thesection.\arabic{subsection}}
\renewcommand\thefigure{\Alph{section}.\arabic{figure}}
\renewcommand\thetable{\Alph{section}.\arabic{table}} 
\begin{appendices}
	\section{Framework Architecture}
	Our framework consists of a global branch and a local branch.
	
	The global branch contains a transformer head and five global blocks. The transformer head is composed of an adaptive global pooling layer and a plain vision transformer block~\cite{dosovitskiy2020image} for global feature extraction. As a result, it is able to process images of arbitrary spatial resolutions at a constant computation cost. Each of the five global blocks is implemented by a fully connected layer followed by a Tanh activation.
	
	The local branch contains a CNN head and a U-shape CNN body. The CNN-head is composed of a single convolution layer followed by "Instance Normalization + ReLU". Meanwhile, the U-shape CNN body has six local convolution blocks: The first and the last convolution blocks are composed of a single convolution followed by a ReLU activation. The remaining four blocks are two down-sampling~(2-th and 3-th conv. blocks) and two up-sampling blocks~(4-th and 5-th conv. blocks), respectively. The detailed structure is illustrated in Table~\ref{table:db} and Table~\ref{table:ub}. The "RCAB" layer refers to the residual channel attention block proposed by~\cite{zhang2018rcan}.
	
	We use a series of mutual guidance blocks~($t_1,t_2,t_3,t_4,t_5$) to fuse the information from a global and local block pair.

	% 	we first utilize a transformer-head and CNN-head for feature extraction. The transformer-head is composed of a adaptive global pooling layer and a plain vision transformer block for global feature extraction. As a result, it is able to process images of arbitrary spatial resolutions at a constant computation cost. Our CNN-head is composed of a single convolution layer followed by ``Instance Normalization + ReLU". Then, at the global branch, it consists of five fully connected layers followed by a ``Tanh" activation for handling global image features. Meanwhile, our local branch body is a U-shape architecture for processing local information. In total, there are six convolution blocks in the local branch body. The first and the last convolution blocks are composed of a single convolution followed by a ReLU activation. The remaining four blocks are two down-sampling~(2-th and 3-th conv. blocks) and up-sampling blocks~(4-th and 5-th conv. blocks). The detailed structure is illustrated in Table~\ref{table:db} and Table~\ref{table:ub}. Each of the five global blocks are connected with a local convolution block with a mutual guidance block~($t_1,t_2,t_3,t_4,t_5$), formulating a global-local interaction block.
	
	\begin{table}[t]
		\begin{center}
			\begin{tabular}{|c|c|c|c}
				\hline 
				Input        	& $ F \in R^{C\times H \times W}$\\ \hline \hline
				Layer1       	& Conv(C,C,3,1,1) + ReLU \\ \hline
				Layer2       	& Conv(C,2C,3,2,1) + ReLU \\ \hline
				Layer3       	& Conv(2C,2C,3,1,1) + ReLU \\ \hline
				Layer4          & RCAB(2C)  \\ \hline
				Layer5          & Conv(2C,2C,1,1,0)  \\ \hline \hline
				Output          & $ F \in R^{2C\times \frac{H}{2} \times \frac{W}{2}}$ \\ \hline
			\end{tabular}
		\end{center}
		\caption{Down-sampling block.}
		\label{table:db}
	\end{table}
	\begin{table}[h]
		\begin{center}
			\begin{tabular}{|c|c|c|c}
				\hline
				Input        	& $ F \in R^{C\times \frac{H}{2} \times \frac{W}{2}}$ \\ \hline \hline
				Layer1       	& Conv(C,4C,3,1,1) + ReLU \\ \hline
				Layer2       	& PixelSuffle(2)\\ \hline
				Layer3       	& Conv(C,C,3,1,1) + ReLU \\ \hline
				Layer4          & RCAB(C) \\ \hline
				Layer5          & Conv(C,C,1,1,0)  \\ \hline \hline
				Output          & $ F \in R^{C\times H \times W}$ \\ \hline
			\end{tabular}
		\end{center}
		\caption{Up-sampling block.}
		\label{table:ub}
	\end{table}
	
	\begin{table}[h]
		\begin{center}
			\begin{tabular}{l|c|c|c|c|c|c } %p{2cm}p{2cm}
				\hline
				Models		&$t_1$         &$t_2$         &$t_3$  &$t_4$ &$t_5$     &Metrics        \\ \hline
				Model-O &           &       	&       	&       	&       	&24.97/0.920          \\	\hline
				Model-A	&\cmark     &       	&       	&       	&       	&25.14/0.921          \\       
				Model-B	&	        & \cmark    &       	&       	&       	&25.19/0.925             \\      
				Model-C	&      		&       	&\cmark     &       	&       	&25.23/0.929             \\  
				Model-D	&	        &     		&     		&\cmark     &       	&25.27/0.928             \\  
				Model-E	&     		&       	&   	 	&       	&\cmark     &25.26/0.928             \\     \hline 
				
			\end{tabular}
		\end{center}
		\caption{The role of different mutual guidance blocks.}
		\label{table:mbt}
	\end{table}
	\begin{table}[h]
		\begin{center}
			\begin{tabular}{l|c|c|c|c|c|c } %p{2cm}p{2cm}
				\hline
				Models		&$t_1$         &$t_2$         &$t_3$  &$t_4$ &$t_5$     &Metrics        \\ \hline
				
				Model-A	&\cmark     &       	&       	&       	&       	&25.14/0.921          \\       
				Model-F	&\cmark	    & \cmark    &  		    &       	&       	&25.31/0.925         \\       
				Model-G	&\cmark	    & \cmark    &\cmark  	&       	&       	&25.35/0.929             \\  
				Model-H	&\cmark	    & \cmark    &\cmark  	&\cmark       	&       &25.43/0.930             \\   
				Model-I	&\cmark	    & \cmark    &\cmark  	&\cmark       	& \cmark& 25.59/0.930            \\ \hline    
			\end{tabular}
		\end{center}
		\caption{Impacts of different mutual guidance block numbers.}
		\label{table:mbn}
	\end{table}
	
	\section{Ablation Study on Mutual Guidance Blocks}
	\noindent{\bf Impact of different Mutual Guidance block.} In this section, we first study the role of different mutual guidance block~($t_1,t_2,t_3,t_4,t_5$). To this end, we train five models~(``A-E"). The $t$-th model only uses the $t$-th mutual guidance block for a one-shot global-local information exchange. In addition, we train another model that ablates all the five mutual guidance blocks ~(named ``Model-O"). As shown in Table~\ref{table:mbt}, compared with Model-O, the other five models~(``A-E") achieve better performance. It verifies the effectiveness of our mutual guidance approach. Also, it can be seen that the deeper mutual fusions~($t_4,t_5$) lead to better performance.  
	
	\noindent{\bf Impact of Mutual Guidance block numbers.} We further check the influence of mutual guidance block numbers. We start from model-A which uses the $t_1$ for global-local information aggregation. Then, we gradually include more mutual guidance blocks and train another four models~(``F-I"). Especially, model-I is our full model that employs all the five mutual guidance blocks for global-local feature fusion. The results are reported in Table~\ref{table:mbn}. It is clear that more mutual guidance blocks achieve higher PSNR values. 
	
	\begin{figure}[t]
		\centering
		\includegraphics[width=1.0\linewidth]{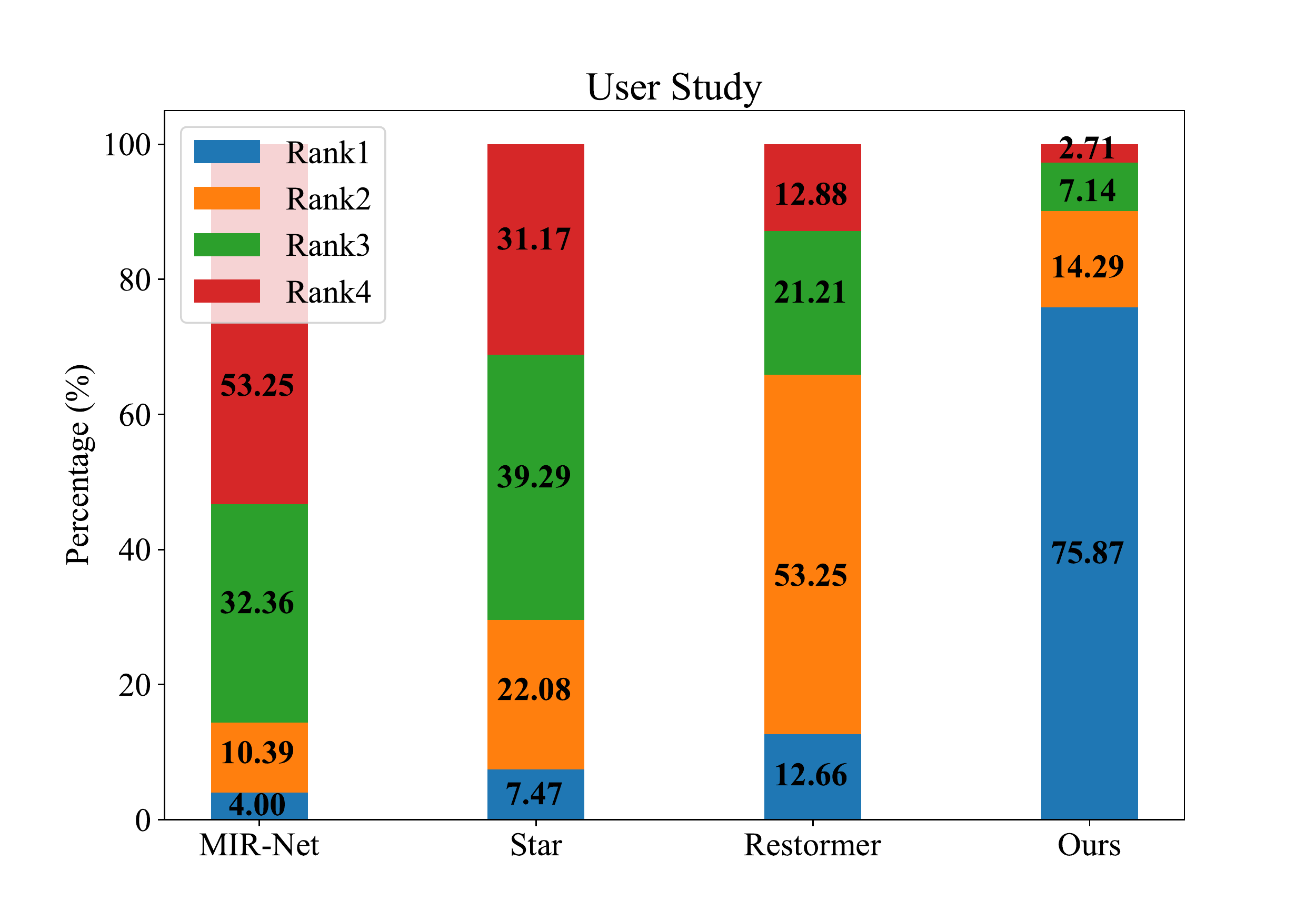} % 
		\caption{The ranking results of the user study. The values on bars represent the percentages of ratings.  }
		\label{fig:user} 
		% mutual_ablation.png
	\end{figure} % 
	
	\section{User Study}
	We further conduct a user study for objective assessment. Apart from our model, we select three representative SOTA methods~(MIR-Net~\cite{Zamir2020MIRNet}, Star~\cite{zhang2021star}, Restormer~\cite{zamir2022restormer}). The first model is a large CNN-based approach and the other two are transformer-based methods. In total, we collected testing images from five sources: (1)  MIT-Adobe-5K-UPE~\cite{Wang_2019_CVPR}, (2) MIT-Adobe-5K-DPE~\cite{chen2018deep}, (3) LOL Synthetic~\cite{yang2021sparse}, (4) LOL Real~\cite{yang2021sparse} and (5) images captured by a mobile phone under various dark environments.
	
	At each time, we randomly select 30 testing cases and ask participants to rank the results produced by the four competing models in terms of global consistency, local contrast, and natural color.   
	In total, 35 participants have finished our questions. The rating results are shown in Figure~\ref{fig:user} . It can be observed that most of our results are rated at the first place~(``Rank1") and our other results still get high rankings.
	\section{Additional Visual Results}
\begin{figure*}[t]
		\centering
		
		\includegraphics[width=1.0\linewidth]{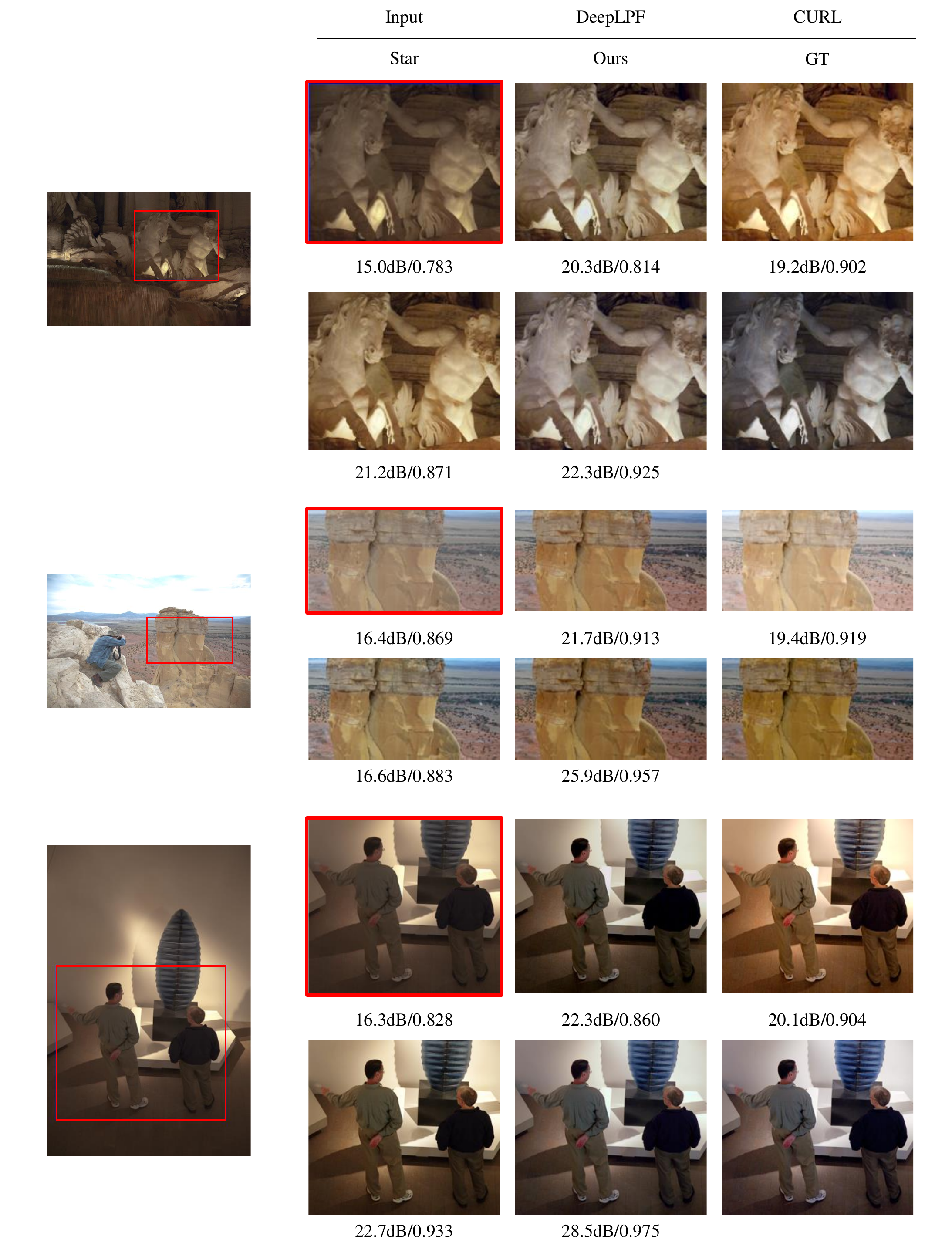} 
		\caption{Qualitative Comparison on Adobe-5K-DPE~\protect\cite{chen2018deep}.  } 
		\label{fig:sota_d1}
	\end{figure*}

\begin{figure*}[t]
		\centering
		
		\includegraphics[width=1.0\linewidth]{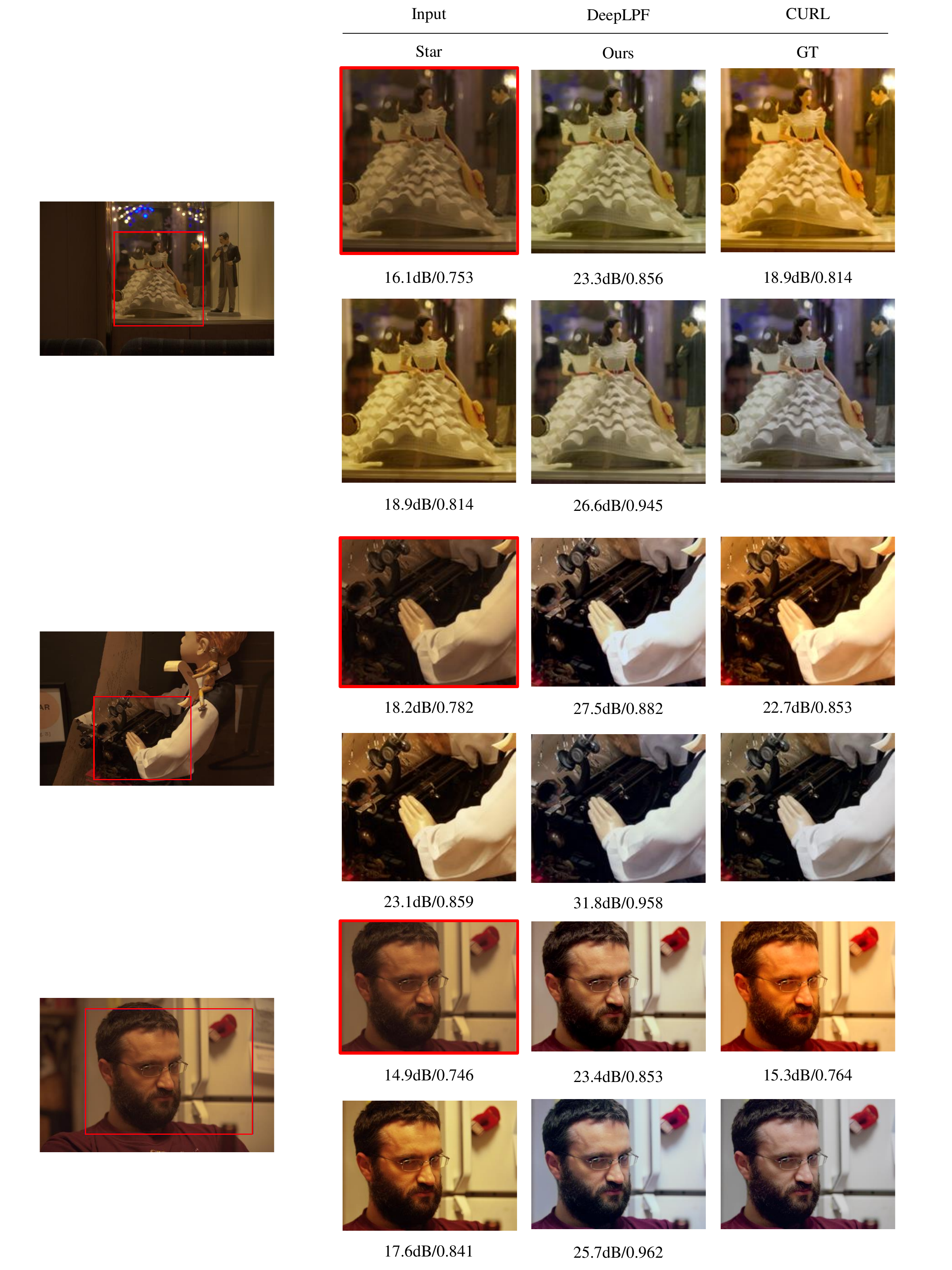} 
		\caption{Qualitative Comparison on Adobe-5K-DPE~\protect\cite{chen2018deep}.  } 
		\label{fig:sota_d2}
	\end{figure*}

\begin{figure*}[t]
		\centering
		
		\includegraphics[width=1.0\linewidth]{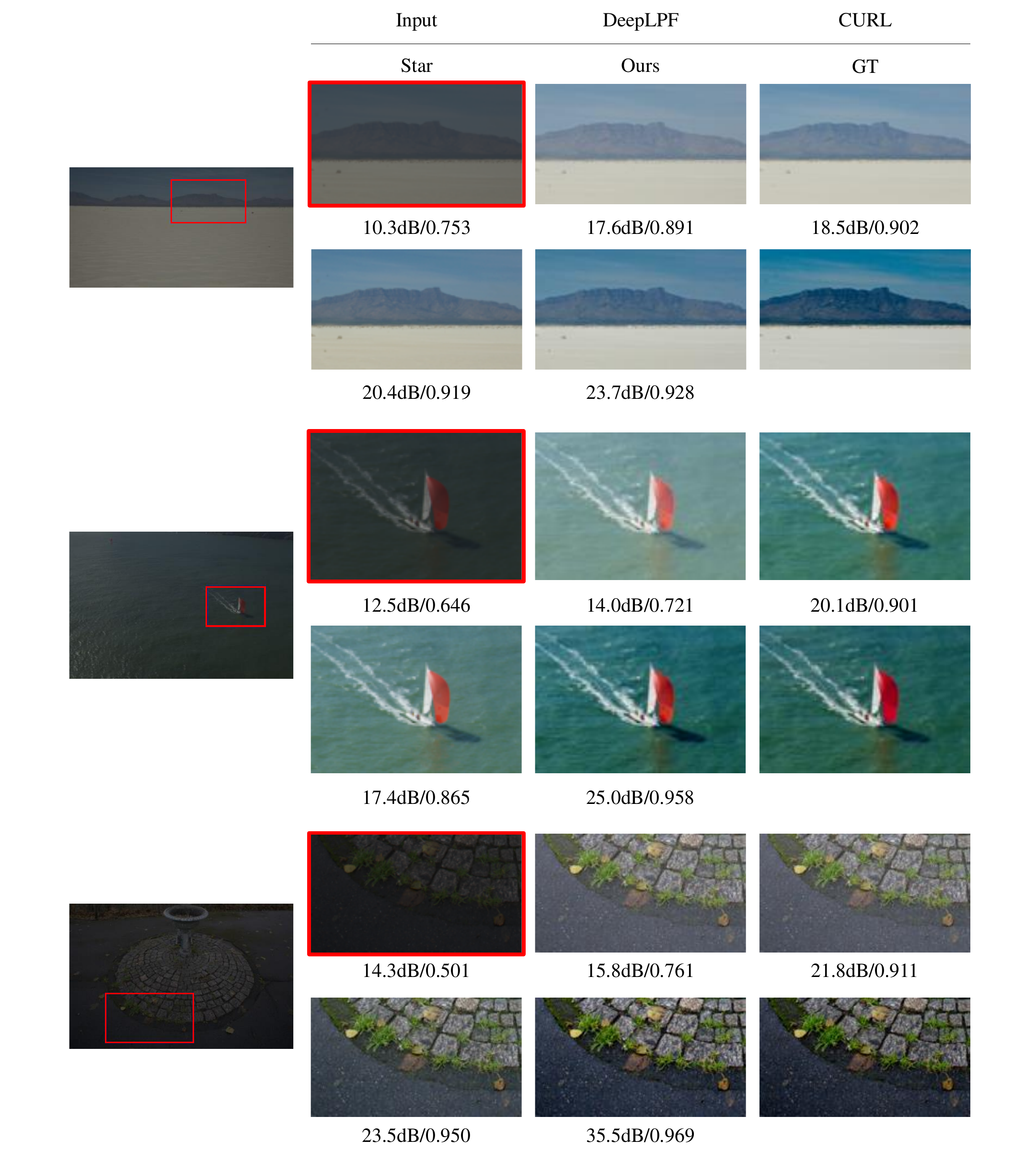} 
		\caption{Qualitative Comparison on Adobe-5K-UP~\protect\cite{Wang_2019_CVPR}.  } 
		\label{fig:sota_u1}
	\end{figure*}

\begin{figure*}[t]
		\centering
		
		\includegraphics[width=1.0\linewidth]{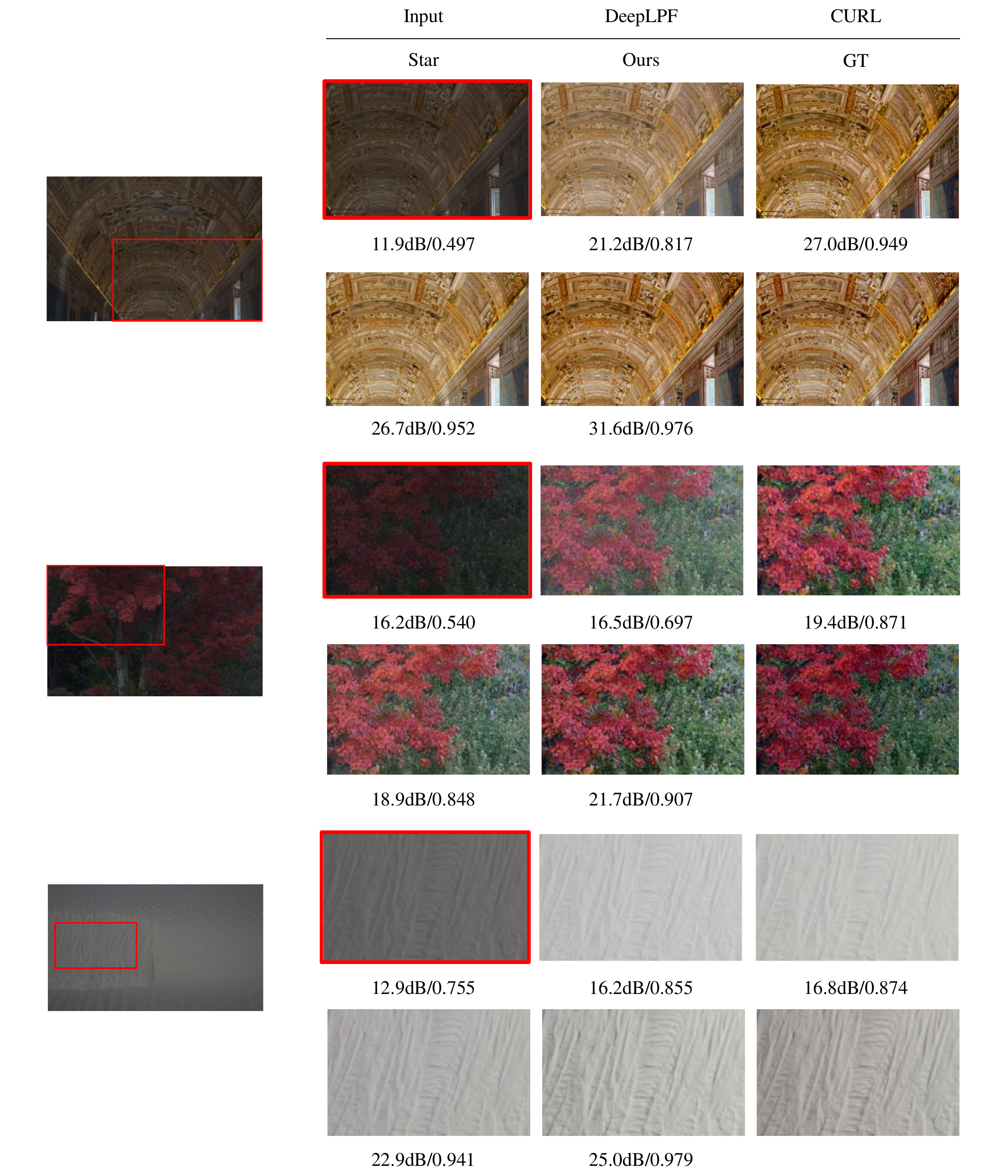} 
		\caption{Qualitative Comparison on Adobe-5K-UPE~\protect\cite{Wang_2019_CVPR}.  } 
		\label{fig:sota_u2}
	\end{figure*}

\begin{figure*}[t]
		\centering
		
		\includegraphics[width=1.0\linewidth]{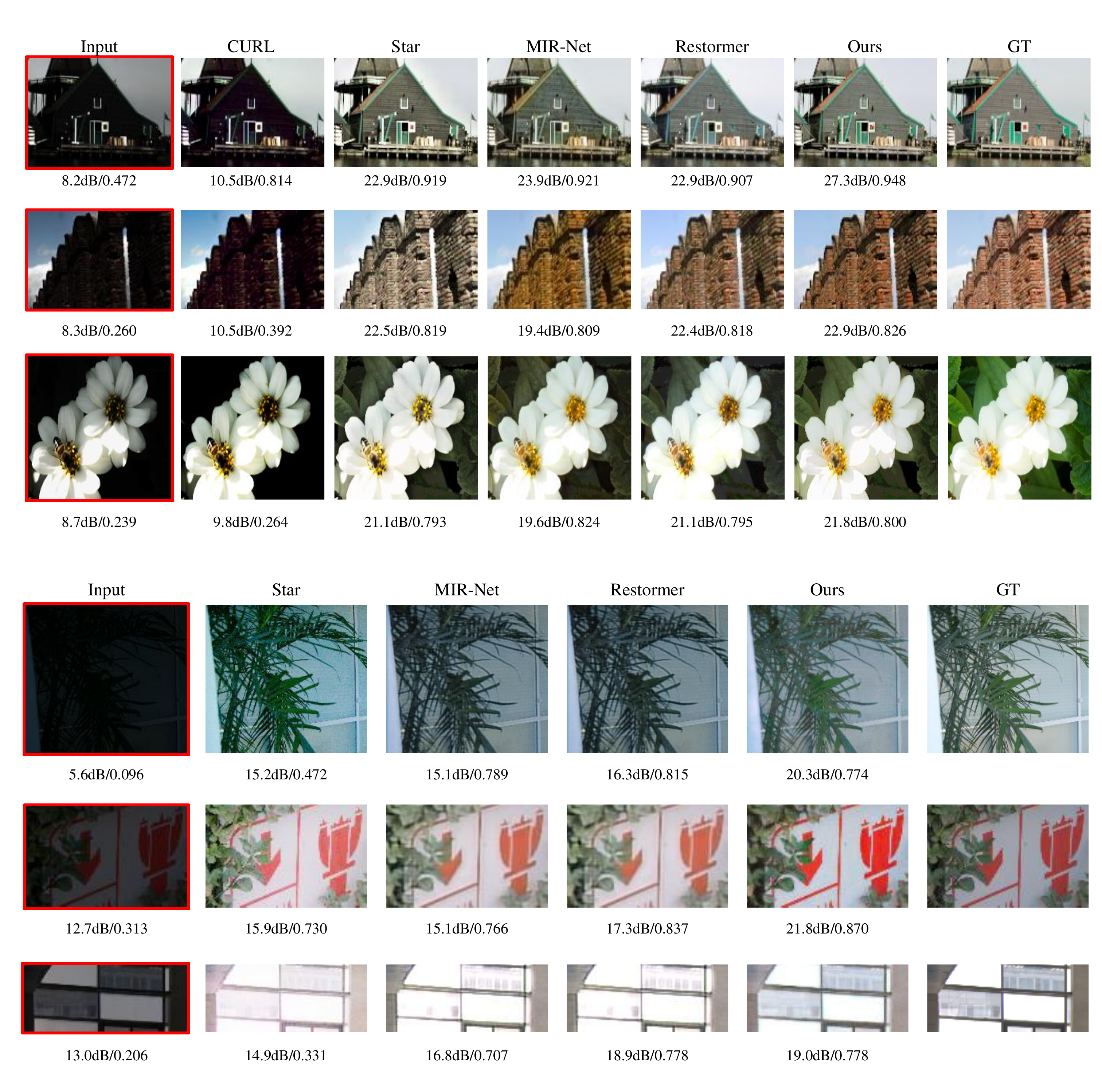} 
		\caption{Qualitative Comparison on LOL-v2 Syntheticl~\protect\cite{yang2021sparse} and LOL-v2 Real~\protect\cite{yang2021sparse}.  } 
		\label{fig:sota_l1}
	\end{figure*}

\begin{figure*}[t]
		\centering
		
		\includegraphics[width=1.0\linewidth]{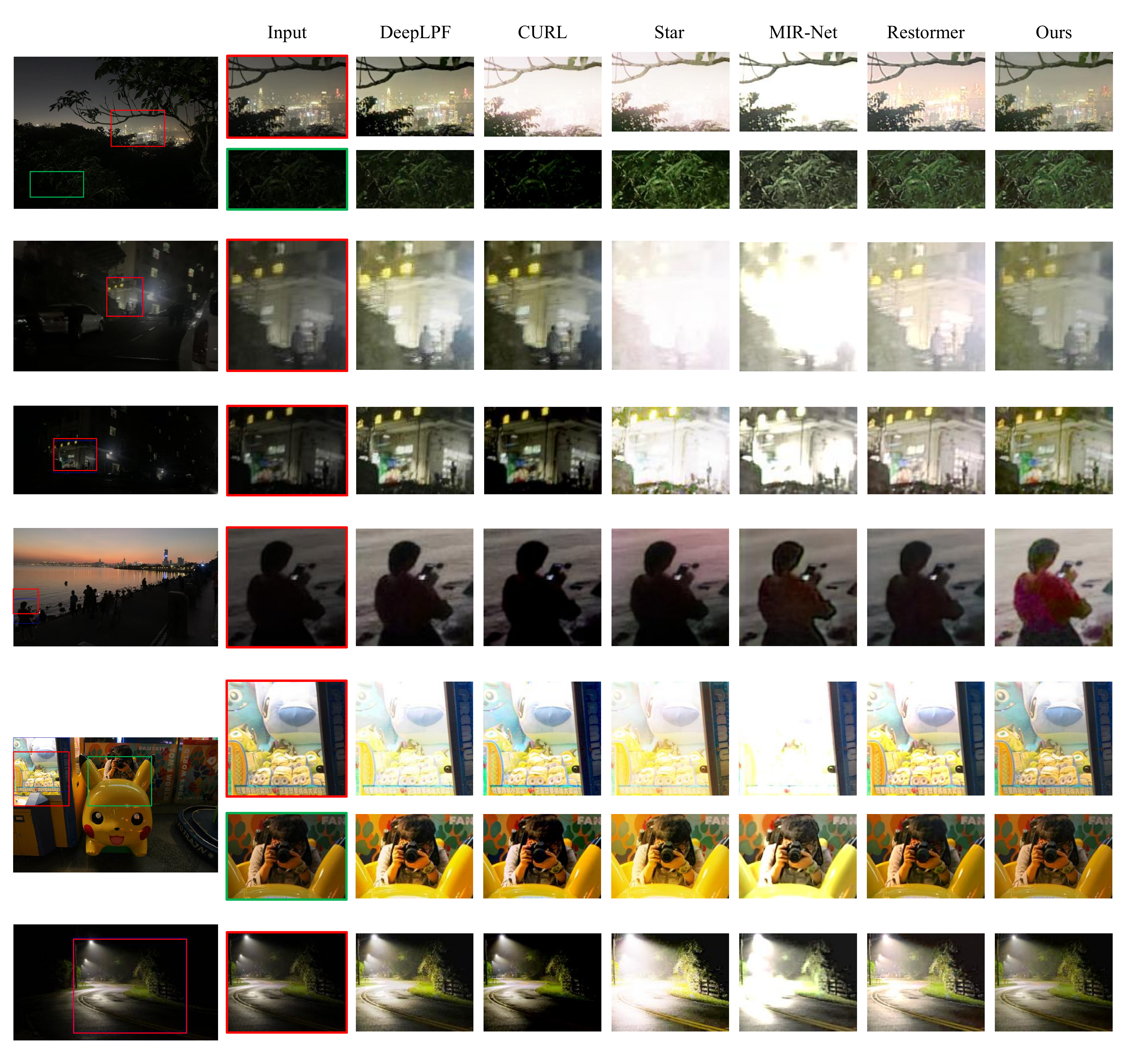} 
		\caption{Several in-the-wild visual examples. The first four images are captured by a mobile phone under various dark environments. And the last two examples are from the Internet.  } 
		\label{fig:sota_w1}
	\end{figure*}

	We present more visual examples from Adobe-5K-UPE~\cite{Wang_2019_CVPR}, MIT-Adobe-5K-DPE~\cite{chen2018deep}, LOL-v2 Real~\cite{yang2021sparse} and real-world images captured by a mobile phone under various dark environments for qualitative analysis. The visual comparison is illustrated in Figure~\ref{fig:sota_u1}~\ref{fig:sota_u2}~\ref{fig:sota_d1}~\ref{fig:sota_d2}~\ref{fig:sota_l1}~\ref{fig:sota_w1}. We make qualitative comparison with SOTA approaches, including DeepLPF~\cite{moran2020deeplpf}, CURL~\cite{moran2021curl}, Star~\cite{zhang2021star}, MIR-Net~\cite{Zamir2020MIRNet}, Restormer~\cite{zamir2022restormer}. It can be observed that our model is capable of enhancing low-quality images with higher PSNR/SSIM values. Especially,
	as shown in Figure~\ref{fig:sota_w1}, the other methods either fail to brighten extremely dark areas with correct object colors or over-enhance the bright contents, resulting in degraded image quality. In contrast, our model recovers more local details and natural image colors, achieving better visibility.
	
	
\end{appendices}
\end{document}